%% file: iclr2027_conference.tex
\documentclass{article} % For LaTeX2e
\usepackage{iclr2027_conference,times}
\usepackage{comment}
\input{math_commands.tex}

\usepackage{hyperref}
\usepackage{url}

\usepackage{titletoc} % for appendix catalogue \startcontents[appendices] + \printcontents[appendices]{}{1}{}

\usepackage[utf8]{inputenc}
\usepackage[T1]{fontenc}

\usepackage{fontawesome5} 

\usepackage{microtype}

\usepackage{inconsolata}

\usepackage{graphicx}
\usepackage{twemojis}
\usepackage{wrapfig} % right-aligned wrapped floats (wrapfigure/wraptable)
\usepackage{capt-of} % captions for side-by-side table/figure minipages

\usepackage{xspace}

\usepackage{xcolor}

\definecolor{darkblue}{rgb}{0, 0, 0.5}
\hypersetup{colorlinks=true, citecolor=darkblue, linkcolor=darkblue, urlcolor=darkblue}

\usepackage{booktabs}
\usepackage{colortbl}
\usepackage{multirow}
\usepackage{makecell}
\usepackage{url}
\usepackage[cachedir=.]{minted}
\usepackage{fancyvrb}
\usepackage{pifont} %-- for ticks
\newcommand{\q}[1]{\textit{``#1''}}
\usepackage{enumitem}  % better itemizations and properties
\usepackage{lipsum} % For dummy text
\usepackage{arydshln} % for dotted lines in table

\newcommand\eat[1]{}
\usepackage{amssymb}  % for \bigstar
\usepackage{amsmath} 
\usepackage{placeins} % for floatbarrier on appendix
\usepackage{tikz}  %https://tex.stackexchange.com/questions/7032/good-way-to-make-textcircled-numbers
\newcommand*\circled[1]{\tikz[baseline=(char.base)]{
            \node[shape=circle,draw,inner sep=0.5pt] (char) {#1};}} 
\definecolor{greenish}{RGB}{102,204,0}
\usepackage{textcomp} % for dollars and cents
\usepackage{xspace}
\usepackage{xcolor}

\definecolor{e}{RGB}{34,139,34}      % green
\definecolor{c}{RGB}{220,20,60}  % red
\definecolor{partial}{RGB}{230,159,0} % amber
\definecolor{nei}{RGB}{12,64,236}          % blue
\definecolor{tableblock}{gray}{0.88}
\newcommand{\cmark}{\textcolor{e}{\ding{51}}}
\newcommand{\xmark}{\textcolor{c}{\ding{55}}}
\DeclareRobustCommand{\pmark}{%
  \tikz[baseline=-0.63ex]{%
    \path[fill=partial,draw=none]
      (0,0.56ex) arc[start angle=90,end angle=270,radius=0.56ex] -- cycle;
    \draw[draw=partial,line width=0.09ex]
      (0,0) circle[radius=0.56ex];
  }%
}
\newcommand{\best}[1]{\textbf{#1}}

\newcommand{\loss}[1]{\,\textcolor{c}{\tiny(-\hspace{0.03em}#1)}}

\newcommand{\threecolgrey}[1]{%
  \rowcolor{tableblock}
  \multicolumn{3}{c}{#1}\\
}

\usepackage[colorinlistoftodos]{todonotes}

\newcommand{\method}{\textsc{R3Con}\xspace}

\title{Realize What Matters: Principled Context \\Representation for Large-Scale Reasoning}

\author{
  \hspace{-0.1cm}Michael Theologitis$^1$\thanks{Equal contribution.}, \, Dean Light$^1$\footnotemark[1], \, Shuyue Stella Li$^1$, \, Benjamin Newman$^1$, 
  \\ 
  \textbf{Yulia Tsvetkov}$^1$\textbf{,} \, \textbf{Dan Suciu}$^1$
  \\
  $^1$University of Washington 
  \\
  \texttt{\{mthe, deanlcs\}@cs.washington.edu}
  \\
  \faGithub\ \url{\methodurl}
}

\newcommand{\methodurl}{https://github.com/michaeltheologitis/r3con}
\newcommand{\evalurl}{https://github.com/michaeltheologitis/r3con-evaluation}

\iclrfinalcopy % Uncomment for camera-ready version, but NOT for submission. 
\begin{document}

\maketitle
\lhead{Preprint}

\begin{abstract}
Solving complex tasks in domains such as science, medicine, law, and finance often requires assembling interdependent information scattered across vast,  heterogeneous  sources far beyond model context limits. Existing approaches tackle this challenge by  organizing information into more manageable representations over which models can reason, such as graphs, textual memories, and retrieval collections. These representations dictate what downstream reasoning is possible and, ultimately, whether it succeeds; yet their design and construction remain largely ad hoc. In this work, drawing on the cognitive theory of relevance realization, we propose concrete principles for designing AI systems that construct effective representations of very large contexts. We analyze existing approaches and show how their successes and failures can be explained by their alignment with these principles, and introduce \method, a harness that systematically operationalizes them. We evaluate \method{} against nine state-of-the-art baselines on two recent benchmarks of reasoning over large document corpora. On these benchmarks, \method{} substantially outperforms the strongest baseline by $20$ and $8.4$ percentage points. It also enables smaller models to outperform much larger ones: \method{} with \texttt{4B} and \texttt{9B} models outperforms all evaluated \texttt{35B} baselines, while \method{} with \texttt{Qwen3.5-35B-A3B} outperforms Claude Code with \texttt{Claude-Sonnet-5} at $3.7\times$ lower cost. Our results show that context representations following our principled approach can reduce reliance on model scale, pointing toward a future of AI systems with frontier-level performance powered by smaller models. 
%Our code is available at \url{\methodurl}.
\end{abstract}

\section{Introduction}\label{sec:intro}

% \vspace{-2mm}
\begin{wrapfigure}[12]{r}{0.5\textwidth}
  \centering
  \vspace{-6.23mm}
  \includegraphics[width=\linewidth]{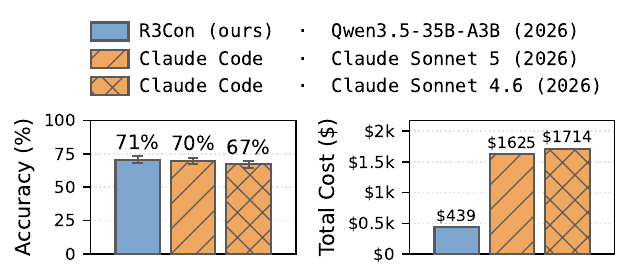}
  \vspace{-7mm}
  \caption{\method{} with a \texttt{3B}-active model outperforms Claude Code equipped with Anthropic's frontier Sonnet models on CorpusQA~\citep{lu2026corpusqa}, while costing roughly $3.7\times$ to $3.9\times$ less.}
  \label{fig:claude-code}
\end{wrapfigure}
Many complex tasks require assembling reasoning from partial, interdependent information sources scattered across  corpora far too large for LLMs to consume directly. Such tasks are common across many domains of human endeavor: from lawyers reconstructing events from evidence~\citep{gay2020onlyway, day2024preparingtrial, taranukhin2026infogatherer}, to scientists synthesizing prior studies~\citep{doi:10.1152/physrev.00028.2022}, financial analysts forecasting from SEC filings~\citep{gibbons2021analyst}, or clinicians assessing patients from diverse medical records~\citep{fan2025supporting}. 

To make such large contexts tractable for reasoning, much existing work first organizes the corpus into more manageable \emph{representations}. In constructing these representations, methods make fundamentally different design decisions. Some pre-commit to sophisticated but fixed structures, such as graphs~\citep{DBLP:conf/icml/GutierrezSQZ025}, retrieval pools~\citep{DBLP:conf/iclr/SarthiATKGM24}, textual memories~\citep{yu2026memagent}, or combinations thereof~\citep{xia2026memoraharmonicmemoryrepresentation}; others choose the structure dynamically~\citep{DBLP:conf/iclr/LiC0L0T0H0L25}. Some construct them offline~\citep{DBLP:journals/corr/abs-2404-16130, DBLP:conf/icml/LeeCFCF24, zhuang2026linearrag}, while others evolve them with the task (e.g., memory agents; see~\citealp{hu2026memoryageaiagents}). However, we lack a principled account of how representations should be constructed in the first place. Each of these many decisions has profound implications for what downstream reasoning is possible (e.g., whether the solution is even accessible), yet it remains unclear which choices are necessary for robust reasoning, and why.

In this work, we provide a principled account of how to construct effective \textsc{\textbf{R}}epresentations of large \textsc{\textbf{Con}}texts, grounded in the cognitive theory of \textsc{\textbf{R}}elevance \textsc{\textbf{R}}ealization~\citep{10.1093/logcom/exp067}. We first introduce this theory, which explains how humans arrive at a problem representation despite working-memory limits, and then translate the underlying cognitive processes into design principles for AI systems~(\S\ref{sec:relevance-realization}). More specifically, such systems should \circled{1} extract relevant context in a \emph{task-specific} and \emph{enactive} manner. Then, they should \circled{2} structure the context using appropriate data structures, with the choice of structure informed by the relevant context and the task. Finally, the relevant context and structure should be brought together to \circled{3}~form the representation. Their interplay makes the representation easier to navigate and, in turn, the solution to the task more readily accessible.

Following these principles, we introduce \textbf{\method{}}~(\S\ref{sec:method}). \method{} uses recursive task-focused summarization to extract the relevant context, and then uses this information to organize the context into dynamically chosen data structures, e.g., tables, nested hierarchies, graphs. The  relevant context and structure form the representation, which \method{} explores and reasons over using a coding agent.

We evaluate \method{} on two recent real-world benchmarks requiring the integration of information across large document corpora, against nine baselines ranging from state-of-the-art representation-construction systems to fully dynamic coding agents (\S\ref{sec:setup}). \textbf{\method{} improves over the strongest baseline by \textcolor{e}{+19.88} percentage points on CorpusQA}~\citep{lu2026corpusqa} \textbf{and \textcolor{e}{+8.42} points on Loong}~\citep{DBLP:conf/emnlp/WangCCL0WYXZLLY24}, as detailed in~\S\ref{sec:results}. Moreover, \method{} achieves the highest accuracy across all domains, including education, finance, real estate, science, and law. \method{} continues to outperform all baselines even with smaller, \textit{cheaper} models. Specifically, \method{} with \texttt{Qwen3.5-4B} on Loong and \texttt{Qwen3.5-9B} on CorpusQA outperforms all baselines using the much larger \texttt{Qwen3.5-35B-A3B}. Finally, we find that \method{} with \texttt{Qwen3.5-35B-A3B} outperforms Claude Code equipped with Anthropic's frontier \texttt{Claude-Sonnet-5} while costing roughly $3.7\times$ less (Fig.~\ref{fig:claude-code}).

Our contributions are threefold: \circled{1} a set of cognitively grounded design principles for constructing effective representations of large contexts; \circled{2} an analysis of existing approaches showing how their failures map onto these principles; and \circled{3} \method{}, an instantiation of these principles that clearly advances the state of the art on information integration tasks.

% \vspace{-3.033mm}
\vspace{-2.033mm}
\section{Motivating Example}\label{sec:motivating-example}

\vspace{-1.633mm}
\begin{wrapfigure}[18]{r}{0.30\textwidth}
  \centering
  \vspace{-5.7mm}
  \includegraphics[width=\linewidth]{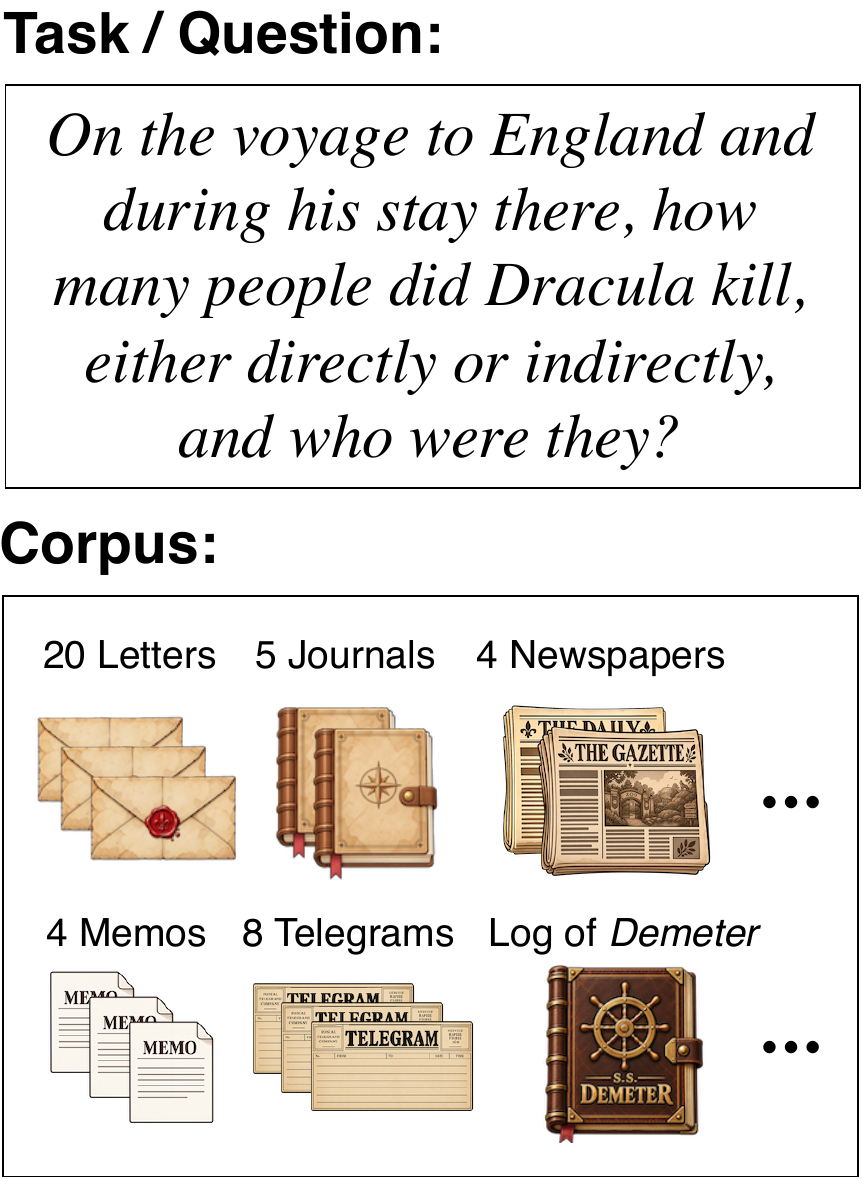}
  \vspace{-6mm}
  \caption{Running example based on Bram Stoker's \textit{Dracula}~\citeyearpar{stoker1897dracula}. Full details in \S\ref{appendix:running-example}.}
  \label{fig:dracula-question}
\end{wrapfigure}

Consider the task in Fig.~\ref{fig:dracula-question} of identifying Dracula's victims in Bram Stoker's novel~\citeyearpar{stoker1897dracula}. The challenge comes from the novel's epistolary form: the story unfolds from a corpus of documents written by its characters---letters, journals, newspaper articles, and others---each recording what its author knew at a particular point in time, often with incomplete information and without understanding the significance of what they observed. Answering correctly therefore requires piecing together information scattered across these incomplete accounts, much like an investigative journalist working through source materials~\citep{Stray14092019, taft2024mississippihospital, spangher-etal-2025-novel}.

\begin{wrapfigure}[17]{r}{0.55\textwidth}
  \centering
  \vspace{-3.0mm}
  \includegraphics[width=\linewidth]{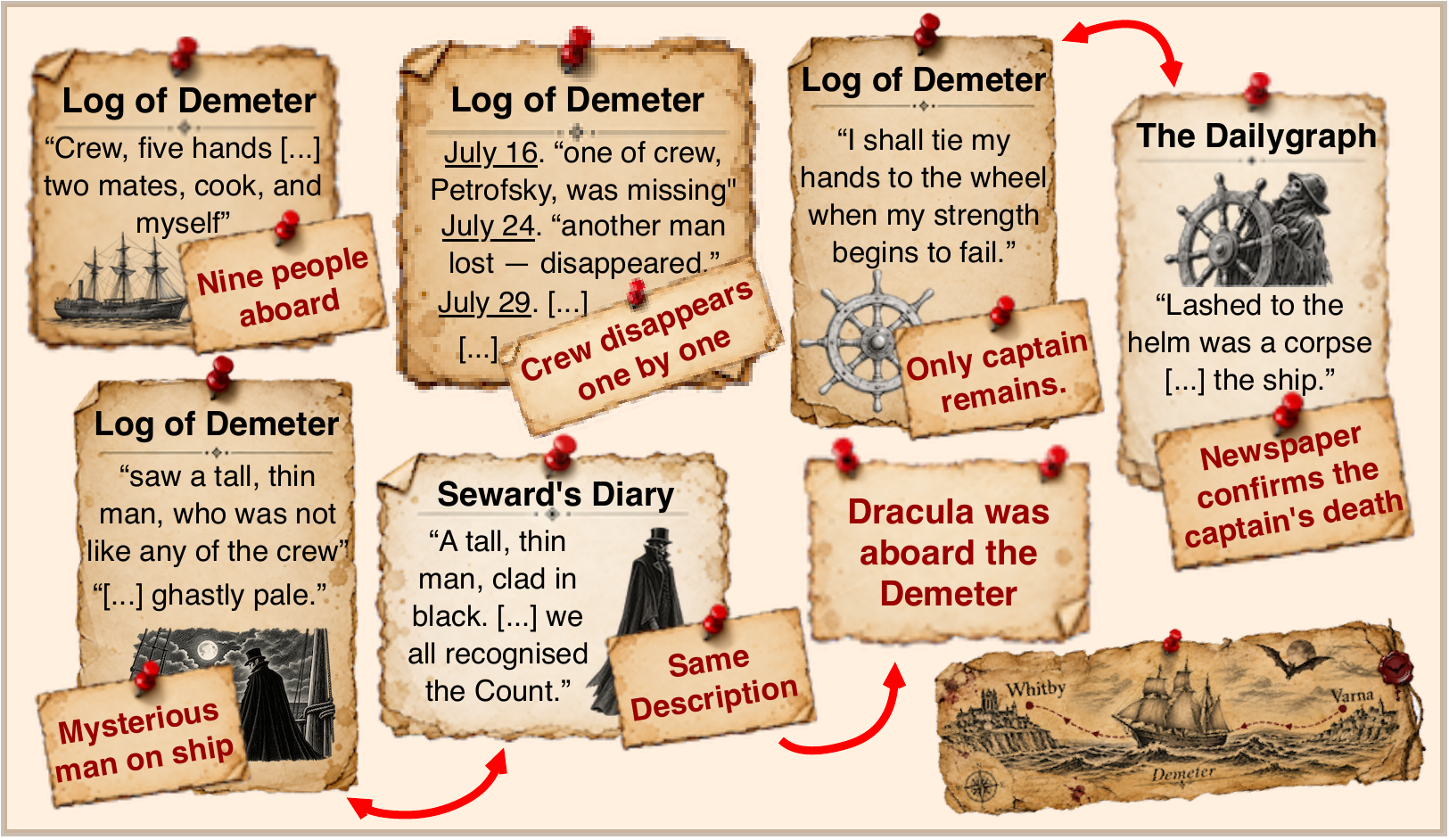}
  \vspace{-6mm}
  \caption{An investigation of the events aboard the ship Demeter. Figuring out that the entire crew of nine died, and that Dracula was behind it, requires connecting evidence across partial and incomplete accounts.}
  \label{fig:demeter-deaths}
\end{wrapfigure}

Take, for example, one part of the story: the tragic events aboard the ship \textit{Demeter} (see Fig.~\ref{fig:demeter-deaths}). On July 6, the ship sets sail for England, with its crew completely unaware that Dracula is hiding among the cargo. The captain establishes in the ship's log that nine people set sail---\q{Crew, five hands ... two mates, cook, and myself}---and then provides scattered reports of the crew disappearing one by one under mysterious circumstances. For instance, on July 16, he writes that \q{one of crew, Petrofsky, was missing}; eight days later, he reports \q{another man lost}; and so on, until only he remains.

After a month at sea, on August~7, the Demeter finally reaches the English coast, in the middle of a violent storm. Two days later, the strange arrival appears in the English newspaper \textit{The Dailygraph}, which also reports an unidentified corpse \q{lashed to the helm.} Taking a closer look at the logs, the captain's last entry notes that he plans to tie his \q{hands to the wheel} until his \q{strength begins to fail.} The newspaper therefore confirms the captain's death, the last of the crew of the Demeter.

\vspace{-1mm}

Connecting all these deaths to Dracula requires piecing together the crew's accounts of a \q{tall, thin man}, \q{ghastly pale} and \q{not like any of the crew} roaming the ship with descriptions found elsewhere in the corpus. For example, Seward's diary later describes the vampire as a \q{tall, thin man, clad in black}. Together with the timeline of events, these accounts reveal that the mysterious man aboard the ship was Dracula, and he was the one responsible for the deaths of its crew (see Fig.~\ref{fig:demeter-deaths}).

All nine baselines we evaluate fail on the Demeter deaths, and therefore fail the overall task in Fig.~\ref{fig:dracula-question}. We later trace these failures back to missing support for context representation  principles (see \S\ref{sec:qualitative-analysis}). For example, HippoRAG2~\citep{DBLP:conf/icml/GutierrezSQZ025} is unable to infer Dracula's involvement because the crew never identify him explicitly, leaving their deaths disconnected from him in the graph. MemAgent~\citep{yu2026memagent} also fails to make this connection: it forgets the deaths from its textual memory before it encounters the context needed to infer that Dracula was aboard (see \S\ref{sec:qualitative-analysis}). We trace both failures to limitations in how these methods interact with the  context (i.e., \textit{enactivity}; \S\ref{sec:cognitive-theory}). Across the whole story, only \method{} manages to identify all 13 victims and hold Dracula fully accountable.

\vspace{-1.233mm}
\section{Relevance Realization}\label{sec:relevance-realization}

\vspace{-0.933mm}
In this section, we introduce the theory of relevance realization~\citep{10.1093/logcom/exp067}, and then translate its underlying cognitive processes into \textit{computational principles for organizing large contexts}.

\vspace{-1.633mm}
\subsection{Cognitive Theory}\label{sec:cognitive-theory}

\vspace{-0.633mm}
Humans routinely reason successfully through information that vastly exceeds our working memory limits~\citep{Baddeley2003Working, Jaeger2024-by}. We do this by making good choices in organizing and relating information, i.e., choosing a problem representation conducive to solving the task. Yet, the space of all possible representations is enormous. In their seminal work, \citet{10.1093/logcom/exp067} put forward the theory of relevance realization, explaining how humans make this otherwise intractable problem tractable by selectively focusing on the relevant information and its relevant structure.

A central observation of \citet{10.1093/logcom/exp067} and \citet{10.1162/jocn_a_01039} is that what is relevant depends on what we are trying to accomplish. The process of realizing relevance is \emph{task-specific}. At the same time, our goals alone are not enough: relevance emerges through continued interaction with the environment, making the process \emph{enactive}\footnote{\textit{Enactive} refers to cognition emerging through continuous, reciprocal interaction between an agent and its environment, through which the agent actively makes sense of its world~\citep{10.7551/mitpress/6730.001.0001, thompson2007mind}.}~\citep{Jaeger2024-by}. Through this interaction, what we learn changes our understanding, which in turn changes what we attend to and how we interpret information. 
As this understanding evolves, we zero in on the relevant information and its relevant structure, which form a representation that makes the problem tractable~\citep{10.1093/logcom/exp067}.

\vspace{-1.533mm}
\subsection{Computational Design Principles Operationalizing the Theory}\label{sec:computational-principles}

\vspace{-1.333mm}
We translate these cognitive processes into design principles for building AI systems capable of constructing effective representations of large contexts. To construct such representations, a system must first \circled{1} \textbf{extract relevant context} in a \textit{task-specific} and \textit{enactive} manner, allowing its understanding of both the task and the context to evolve. It should then \circled{2} \textbf{structure the context} guided by the extracted relevance and the task in a \textit{flexible} way: the system should be free to choose both the structure's semantic organization (e.g., entities and relations, claims and evidence, etc.) and its data structure (e.g., tables, graphs, hierarchies, lists, etc.). Together, the extracted context and resulting structure \circled{3} \textbf{form the representation} that the system can then explore and reason over. Their interplay is important: the relevant information helps navigating the structure (i.e., contextualizes it), and the structure makes reasoning over relevant information easier. 

\begin{figure}[t]
  \centering
  \includegraphics[width=0.96\linewidth]{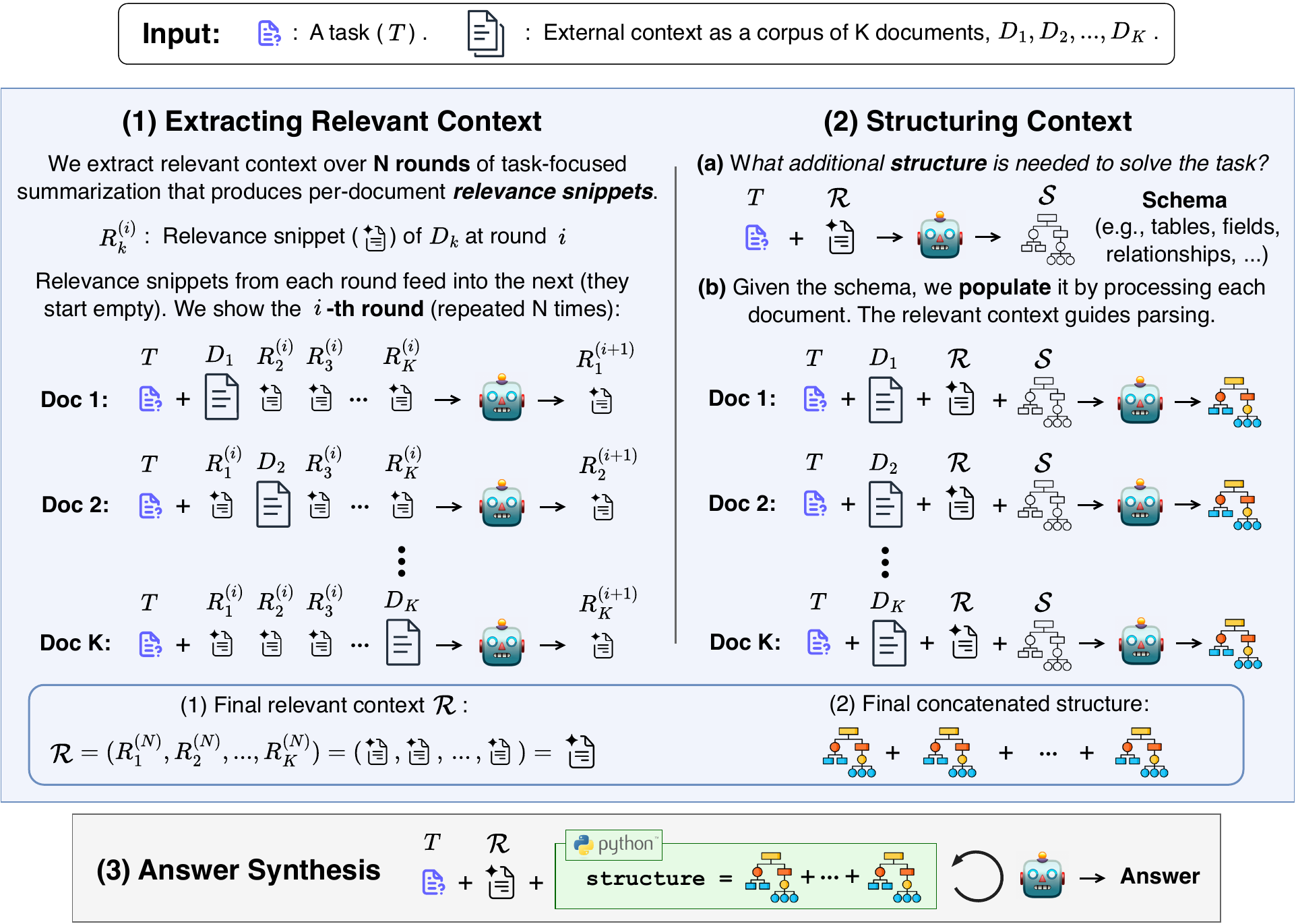}
  \vspace{-2mm}
  \caption{\textbf{\method{}} answers a task over a large context by first constructing an effective representation using our design principles (\S\ref{sec:computational-principles}). It  extracts the relevant context $\mathcal{R}$ (Phase 1) and then uses $\mathcal{R}$ to further structure the context (Phase 2). Together, the extracted relevant context and structured context form the representation, which is then used by a coding agent to produce the answer (Phase 3).}
  \label{fig:r3con}
\end{figure}

\vspace{-2.933mm}
\section{\method}\label{sec:method}

\vspace{-1.933mm}
In this section, we introduce \textbf{\method{}} which instantiates the principles derived from the theory of \textsc{\textbf{R}}elevance \textsc{\textbf{R}}ealization for constructing an effective \textsc{\textbf{R}}epresentation of large \textsc{\textbf{Con}}texts (\S\ref{sec:computational-principles}).

\noindent \textbf{Problem Formulation.} Let $T$ be a task over a large context $\mathcal{D} = \{D_1, \ldots, D_K\}$, represented as a corpus of $K$ text documents. The goal is to answer $T$ using information available in the context. 

\vspace{-2.333mm}
\subsection{Extracting Relevant Context}\label{sec:extracting-relevance}

\vspace{-1.333mm}
In a sea of available information, the foremost challenge is extracting the context relevant to the task. Relevance, however, is not an intrinsic property of a passage or document: information that appears irrelevant in isolation may become crucial once we interact and consider other context.

Recall from our running example that \textit{The Dailygraph} reports an unidentified corpse \q{lashed to the helm} (see Fig.~\ref{fig:demeter-deaths}). In isolation, this appears to be a tragic newspaper story and the relevance of these details to our task remains unclear. However, the final entry of the log of Demeter states the captain's plan to tie his \q{hands to the wheel} until his \q{strength begins to fail.} Revisiting the newspaper account with this new context changes its interpretation and relevance: the unidentified corpse can now be recognized as the captain, and thus as Dracula's last victim on the voyage to England.

This illustrates the enactive nature of extracting relevance discussed in \S\ref{sec:relevance-realization}: as we interact with the broader environment (i.e., corpus), newly acquired context can change the interpretation---and therefore the relevance---of information encountered earlier. Thus, we \emph{extract relevance} iteratively, revisiting information as our understanding of each document and of the corpus as a whole evolves.

\noindent \textbf{Instantiation Approach.} More specifically, we revisit each document over $N$ rounds using task-focused summarization~\citep{Dang2005OverviewOD}. At each round $i$, for each document $D_k$, we construct a \emph{relevance snippet} $R_k^{(i)}$. As illustrated in Phase 1 of Fig.~\ref{fig:r3con}, the language model receives the query $T$, the original document $D_k$, and the snippets of all \emph{other} documents from the previous round (starting with $R_k^{(0)}=\emptyset$ for all $k$). It is then asked to summarize what the document contributes toward the task, retaining only task-relevant information (see Fig.~\ref{fig:summarization_prompt} for the prompt). Here, we must be careful: what is relevant for the task might not be obvious in the current round---for example, necessary multi-hop connections may not yet have been discovered, as in the newspaper story about the unidentified dead helmsman. Therefore, when constructing $D_k$'s relevance snippet at each round, we exclude its own previous state as it is potentially incomplete and distracting. This allows the model to re-interpret the document from scratch in light of the current corpus-wide understanding. Finally, after $N$ rounds, we retain the final \textit{relevant context} $\mathcal{R} = (R_1^{(N)}, R_2^{(N)}, ..., R_K^{(N)})$; we omit $N$ from $\mathcal{R}$ for simplicity.

% Figure~\ref{fig:relevance-snippets} illustrates this evolution across two rounds in our running example. Initially, each document provides only a limited view. By the second round, the broader context allows the necessary causal connections to emerge and relevance to be realized.

\begin{figure}[tbp]
  \centering
  \includegraphics[width=0.97\linewidth]{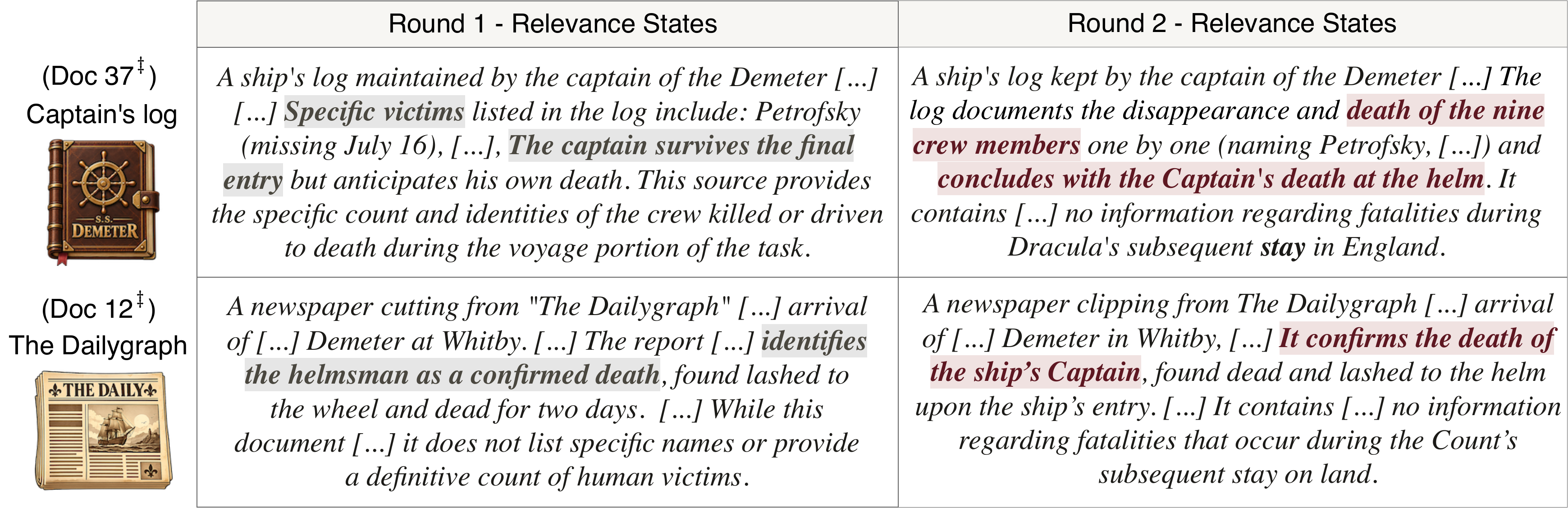}
  \vspace{-2mm}
  \caption{Evolution of the relevance snippets for the ship's log ($D_{37}$) and The Dailygraph newspaper ($D_{12}$) across two rounds of extracting relevance. For example, the top-left and top-right cells show $R_{37}^{(1)}$ and $R_{37}^{(2)}$, respectively. $^\ddagger$Document ordering is arbitrary. The full snippets are shown in Fig.~\ref{fig:full-relevance-snippets}.}
  \label{fig:relevance-snippets}
\end{figure}

Fig.~\ref{fig:relevance-snippets} illustrates this evolution across two rounds in our running example: by the second round, the broader context allows the necessary causal connections to emerge and relevance to be realized.

\vspace{-1.933mm}
\subsection{Structuring}\label{sec:structuring}

\vspace{-1.333mm}
Now that we have extracted the relevant context, we turn our focus to structuring it. We use $\mathcal{R}$ together with the task to guide both the choice of structure and how it is populated.

\noindent\textbf{Schema.} Given the task $T$ and $\mathcal{R}$, the model is asked to propose a schema $S$ that fully specifies how the information should be organized and related, including what information to represent, its attributes, relationships, nesting, and data types (see Fig.~\ref{fig:proposed_prompt} for the prompt). For example, for the Dracula task, the model proposes a schema that organizes death events into two groups---those that happened during the voyage and those during Dracula's stay in England---and records attributes such as the name of the victim, cause of death, and supporting evidence (see Fig.~\ref{fig:dracula-pydantic-schema} in \S\ref{appendix:examples-structures}). For other tasks, the schema can take more complex forms. For example, financial tasks may call for hierarchical or graph-like organizations of attributes (see Fig.~\ref{fig:corpusqa-cash-flow-schema} for one such chosen structure).

\noindent\textbf{Parsing.} Finally, given the schema $S$, we return to the original documents and populate it. Each document $D_k$ is processed independently with the task $T$, the schema, and the relevant context $\mathcal{R}$ as context (see Fig.~\ref{fig:r3con}; the prompt is in Fig.~\ref{fig:parser_prompt}). Crucially, $\mathcal{R}$ provides the contextual understanding needed to resolve ambiguities that commonly arise when structuring text~\citep{martinelli-etal-2025-xcore}.

\vspace{-1.933mm}
\subsection{Answer Synthesis}\label{sec:reasoning}

\vspace{-1.333mm}
Together, the previous two stages construct the representation: the relevant context $\mathcal{R}$ and the resulting structure (\S\ref{sec:extracting-relevance} and \S\ref{sec:structuring}). We use a standard agentic coding loop to reason over the representation~\citep{DBLP:journals/corr/abs-2512-24601}: this allows the model to explore and manipulate the structured data programmatically without having to fit the entire structure into its prompt. We provide $\mathcal{R}$ directly in the prompt (Fig.~\ref{fig:codeact_prompt}) and place the resulting structured data inside a Python Read-Eval-Print Loop (REPL) environment (see Fig.~\ref{fig:r3con}). We also study giving the full representation in-context in \S\ref{appendix:downsteam-reasoning}.

\vspace{-1.633mm}
\section{Experimental Setup}\label{sec:setup}

\vspace{-1.633mm}
\subsection{Benchmarks}

\vspace{-0.933mm}
We select benchmarks that test reasoning over interdependent information distributed across large document collections from diverse real-world domains, including finance, education, real estate, science, and law. We evaluate on two complementary benchmarks. First, \textbf{CorpusQA}~\citep{lu2026corpusqa} focuses on scale and dispersion, with questions requiring the integration of information from hundreds of interconnected real-world documents at a time. Second,  \textbf{Loong}~\citep{DBLP:conf/emnlp/WangCCL0WYXZLLY24} offers  diverse forms of reasoning across a collection of documents, with each question categorized by the reasoning it requires. This allows us to analyze performance separately on locating, comparing, clustering, and chaining evidence across the long  context.

\noindent\textbf{Metrics.} Both benchmarks report accuracy and use an LLM-as-a-judge. We use the original prompts, with \texttt{gpt-5.4-mini}~\citep{openai2026gpt54mininano} as the judge. Further details are provided in Appendix~\ref{appendix:benchmarks}.

\vspace{-1.633mm}
\subsection{Baselines}

\vspace{-0.933mm}
Reasoning over large contexts has been approached in several different ways. We evaluate state-of-the-art methods from different lines of work (see \S\ref{appendix:baselines} for details).

First, \emph{representation-based approaches} rely on transforming the context into more manageable representations that can then be used for reasoning. \mbox{\textbf{HippoRAG2}}~\citep{DBLP:conf/icml/GutierrezSQZ025} and \mbox{\textbf{LinearRAG}}~\citep{zhuang2026linearrag} build sophisticated graphs, while \textbf{StructRAG}~\citep{DBLP:conf/iclr/LiC0L0T0H0L25} chooses the structure dynamically from several options. Other work uses textual representations, as in \mbox{\textbf{RAPTOR}}~\citep{DBLP:conf/iclr/SarthiATKGM24}, which builds a hierarchy of increasingly abstract summaries used for retrieval, and memory agents such as \textbf{ReadAgent}~\citep{DBLP:conf/icml/LeeCFCF24} and \textbf{MemAgent}~\citep{yu2026memagent}.

On the other hand, \emph{coding- and tool-calling agents} allow the model to dynamically adapt its reasoning strategy through multi-step loops based on the task and interactions with the context. We include \textbf{Claude Code}~\citep{ClaudeCode}, \textbf{CodeAgent}~\citep{smolagents}, and \textbf{Recursive Language Models (RLMs)}~\citep{DBLP:journals/corr/abs-2512-24601}, which allow models to recursively call themselves within a sandbox, enabling powerful decomposition patterns. Finally, we include \textbf{A-RAG}~\citep{DBLP:journals/corr/abs-2602-03442}, which equips the agent with multiple specialized retrieval tools.

\vspace{-1.633mm}
\subsection{Models}

\vspace{-0.933mm}
We use \texttt{Qwen3.5-35B-A3B}~\citep{qwen3.5} as the default model across \method{} and all baselines, unless otherwise specified. We additionally evaluate \texttt{Qwen3.5-4B} and \texttt{Qwen3.5-9B} with \method{} and the strongest baselines. For MemAgent~\citep{yu2026memagent}, we use its RL-trained \texttt{RL-MemoryAgent-14B}. For experiments with Claude Code, we use \texttt{Claude-Sonnet-4.6}~\citep{ClaudeSonnet46} with \texttt{xhigh} reasoning and \texttt{Claude-Sonnet-5}~\citep{ClaudeSonnet5} with its default \texttt{high} reasoning. Finally, we use OpenAI's \texttt{text-embedding-3-small} for all embedding calls across baselines.

\noindent\textbf{Cost.} We report cost in USD (\$) to provide a common unit across methods, including the cost of embedding models, using the lowest listed prices on OpenRouter\footnote{\url{https://openrouter.ai/models}} at the time of writing.

\vspace{-1.633mm}
\subsection{Implementation Details}

\vspace{-0.933mm}
For \method{}, we use two rounds of extracting relevance ($N=2$) throughout, unless otherwise specified. We analyze the effect of the number of rounds in \S\ref{appendix:number-rounds}. Further implementation details for each baseline are provided in \S\ref{appendix:baselines}. Infrastructure details are provided in \S\ref{sec:infrastcure}.

\vspace{-1.633mm}
\section{Results}\label{sec:results}

\vspace{-1.633mm}
\subsection{Overall Performance and Cost}\label{sec:overall-performance-and-cost}

\vspace{-0.533mm}
\begin{table*}[t]
\caption{Accuracy (\% $\uparrow$) across real-world domains requiring reasoning over large document corpora. \method{} achieves the highest accuracy across all domains, outperforming the strongest baseline by \textcolor{e}{+19.88} pp on CorpusQA and \textcolor{e}{+8.42} pp on Loong overall (SEs in Tab.~\ref{tab:overall-results-error-bars}). Accuracy is micro-averaged.}
\vspace{-3mm}
\label{tab:overall-results}
\begin{center}
\resizebox{\textwidth}{!}{%
\begin{tabular}{l@{\hspace{8pt}} *{8}{c}}
\toprule
\multirow{2}{*}{\textbf{Method}}
& \multicolumn{4}{c}{\textbf{CorpusQA}~\citep{lu2026corpusqa}}
& \multicolumn{4}{c}{\textbf{Loong}~\citep{DBLP:conf/emnlp/WangCCL0WYXZLLY24}} \\
\cmidrule(lr){2-5}
\cmidrule(lr){6-9}
& \textbf{Edu.} & \textbf{Finance} & \textbf{Real-estate} & \textbf{Overall}
& \textbf{Paper} & \textbf{Finance} & \textbf{Legal} & \textbf{Overall} \\
\midrule
RAPTOR~\citeyearpar{DBLP:conf/iclr/SarthiATKGM24} & 8.00 & 11.11 & 17.95 & 12.06 & 0.00 & 22.17 & 5.71 & 12.12 \\
ReadAgent~\citeyearpar{DBLP:conf/icml/LeeCFCF24} & 42.68 & 37.42 & 31.25 & 37.23 & 15.15 & 35.33 & 14.93 & 24.49 \\
MemAgent~\citeyearpar{yu2026memagent} & 7.32 & 7.83 & 8.64 & 7.90 & 0.00 & 26.80 & 4.00 & 13.63 \\
HippoRAG2~\citeyearpar{DBLP:conf/icml/GutierrezSQZ025} & 7.32 & 7.83 & 20.99 & 10.94 & 0.32 & 17.42 & 6.29 & 10.06 \\
LinearRAG~\citeyearpar{zhuang2026linearrag} & 8.54 & 6.63 & 16.05 & 9.42 & 0.61 & 9.38 & 5.43 & 6.04 \\
StructRAG~\citeyearpar{DBLP:conf/iclr/LiC0L0T0H0L25} & 53.66 & 55.42 & 38.27 & 50.76 & 19.70 & 32.66 & 12.32 & 23.72 \\
CodeAgent~\citeyearpar{smolagents} & 14.63 & 26.99 & 26.25 & 23.69 & 13.80 & 40.07 & 24.26 & 29.01 \\
RLM~\citeyearpar{DBLP:journals/corr/abs-2512-24601} & 13.41 & 37.58 & 20.51 & 27.38 & 12.77 & 41.67 & 22.77 & 28.96 \\
A-RAG~\citeyearpar{DBLP:journals/corr/abs-2602-03442} & 8.54 & 7.23 & 18.52 & 10.33 & 0.61 & 18.43 & 2.86 & 9.55 \\
\method{} (ours) & \best{75.61} & \best{82.42} & \best{41.25} & \best{70.64} & \best{27.58} & \best{50.59} & \best{24.29} & \best{37.43} \\
\bottomrule
\end{tabular}
}
\end{center}
\vspace{+2mm}
\begin{minipage}{\textwidth}
  \centering
  \vspace{-1.5mm}
  \includegraphics[width=0.99\linewidth]{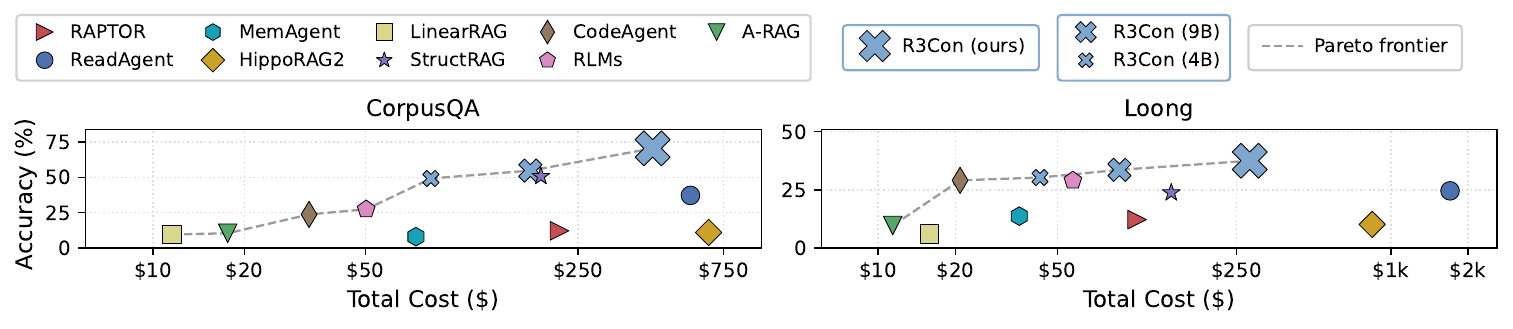}
  \vspace{-3.5mm}
  \captionof{figure}{\method{} lies on the Pareto frontier of the \textit{overall accuracy vs. cost} trade-off across model sizes. Smaller \method{} markers show runs with smaller models from the same family (\texttt{4B} and \texttt{9B}).}
  \label{fig:cross_cost}
\end{minipage}
\end{table*}

\vspace{-0.933mm}
\noindent\textbf{\method{} outperforms baselines by large margins across benchmarks and domains.} \method{} improves over the strongest baseline by \textcolor{e}{+19.88} percentage points on CorpusQA and \textcolor{e}{+8.42} points on Loong (Tab.~\ref{tab:overall-results}). Moreover, \method{} achieves the highest accuracy across all evaluated domains, including education, finance, real estate, academic papers and legal cases (Tab.~\ref{tab:overall-results}). On CorpusQA, improvements over the strongest baseline reach \textcolor{e}{+27.00} pp in finance and \textcolor{e}{+21.95} pp in education, while on Loong they reach \textcolor{e}{+7.88} pp on academic papers and \textcolor{e}{+8.92} pp in financial reports.

\noindent\textbf{\method{} continues to outperform all baselines even with smaller, \textit{cheaper} models.} With the default \texttt{Qwen3.5-35B-A3B} used across \method{} and all baselines, \method{} is on the higher end of cost, although it remains less expensive than ReadAgent and HippoRAG2 (see Fig.~\ref{fig:cross_cost}). Importantly, we can  reduce this cost by using smaller models while retaining superior accuracy over all baselines (see Fig.~\ref{fig:cross_cost}). More specifically, on Loong, \method{} with \texttt{Qwen3.5-4B} achieves 30.23\% accuracy, surpassing all baselines using the default \texttt{35B} model, while reducing its cost by $6.58\times$. The same story holds on CorpusQA, where \method{} with \texttt{Qwen3.5-9B} achieves 54.88\% accuracy, surpassing all baselines with $2.52\times$ lower cost. Notably, the strongest baselines become substantially less competitive when using these smaller models (see \S\ref{appendix:different-models}). Finally, beyond using smaller models, cost can be further reduced by decreasing the number of rounds $N$ with little loss in accuracy (see \S\ref{appendix:number-rounds}).

\vspace{-2.033mm}
\subsection{Fine-Grained Analysis of Reasoning Facets}

\vspace{-0.933mm}
We now investigate how \method{}'s performance varies across types of reasoning using Loong~\citeyearpar{DBLP:conf/emnlp/WangCCL0WYXZLLY24}, since each example is annotated with the reasoning facet required to answer it, as shown in Fig.~\ref{fig:reasoning_tasks}.

\noindent\textbf{\method{} improves across all reasoning categories.} \method{} achieves the highest accuracy in every category, outperforming the strongest baseline by \textcolor{e}{+8.42} percentage points overall (Tab.~\ref{tab:context-results}).

\noindent\textbf{\method{}'s improvements increase with reasoning difficulty.} Based on overall performance across methods, ``chain of reasoning'' and ``clustering'' are the most difficult facets (Tab.~\ref{tab:context-results}), and also where \method{} has the largest advantage over the strongest baseline. Its gains nearly double from \textcolor{e}{+2.45} and \textcolor{e}{+4.74} percentage points on ``spotlight locating'' and ``comparison'' to \textcolor{e}{+8.89} and \textcolor{e}{+8.03} on ``chain of reasoning'' and ``clustering'', respectively.

\newcommand{\reasoningtablewidth}{0.56\textwidth}
\newcommand{\reasoningfigurewidth}{0.41\textwidth}
\begin{figure*}[t]
\centering
\begin{minipage}[t]{\reasoningtablewidth}
\centering
\vspace{-0pt}
\captionof{table}{Accuracy (\% $\uparrow$) across Loong's reasoning facets.}
\label{tab:context-results}
\vspace{+5pt}
\resizebox{\linewidth}{!}{%
\begin{tabular}{l@{\hspace{8pt}} *{5}{c}}
\toprule
\textbf{Method}
& \textbf{Spot.}
& \textbf{Comp.}
& \textbf{Chain}
& \textbf{Clust.}
& \textbf{Overall} \\
\midrule
RAPTOR~\citeyearpar{DBLP:conf/iclr/SarthiATKGM24} & 51.31 & 16.25 & 2.04 & 1.54 & 12.12 \\
ReadAgent~\citeyearpar{DBLP:conf/icml/LeeCFCF24} & 60.11 & 33.77 & 18.49 & 10.70 & 24.49 \\
MemAgent~\citeyearpar{yu2026memagent} & 42.13 & 35.00 & 0.64 & 0.95 & 13.63 \\
HippoRAG2~\citeyearpar{DBLP:conf/icml/GutierrezSQZ025} & 46.70 & 13.33 & 0.33 & 0.38 & 10.06 \\
LinearRAG~\citeyearpar{zhuang2026linearrag} & 32.49 & 3.75 & 0.64 & 0.38 & 6.04 \\
StructRAG~\citeyearpar{DBLP:conf/iclr/LiC0L0T0H0L25} & 44.16 & 30.83 & 24.76 & 12.19 & 23.72 \\
CodeAgent~\citeyearpar{smolagents} & 63.54 & 45.26 & 22.80 & 12.90 & 29.01 \\
RLM~\citeyearpar{DBLP:journals/corr/abs-2512-24601} & 61.14 & 45.15 & 19.68 & 15.27 & 28.96 \\
A-RAG~\citeyearpar{DBLP:journals/corr/abs-2602-03442} & 35.03 & 14.58 & 1.28 & 2.65 & 9.55 \\
\method{} (ours) & \best{65.99} & \best{50.00} & \best{33.65} & \best{23.30} & \best{37.43} \\
% \hspace{1em}\textit{vs. best baseline}
% & \footnotesize{\textcolor{e}{+4.06 pp}}
% & \footnotesize{\textcolor{e}{+4.85 pp}}
% & \footnotesize{\textcolor{e}{+8.89 pp}}
% & \footnotesize{\textcolor{e}{+8.03 pp}}
% & \footnotesize{\textcolor{e}{+8.42 pp}} \\
%
%
%
%
%
%
%
%
%
%
%
%
%
\bottomrule
\end{tabular}
}
\end{minipage}
\hfill
\begin{minipage}[t]{\reasoningfigurewidth}
\vspace{+12pt}
\centering
\includegraphics[width=\linewidth]{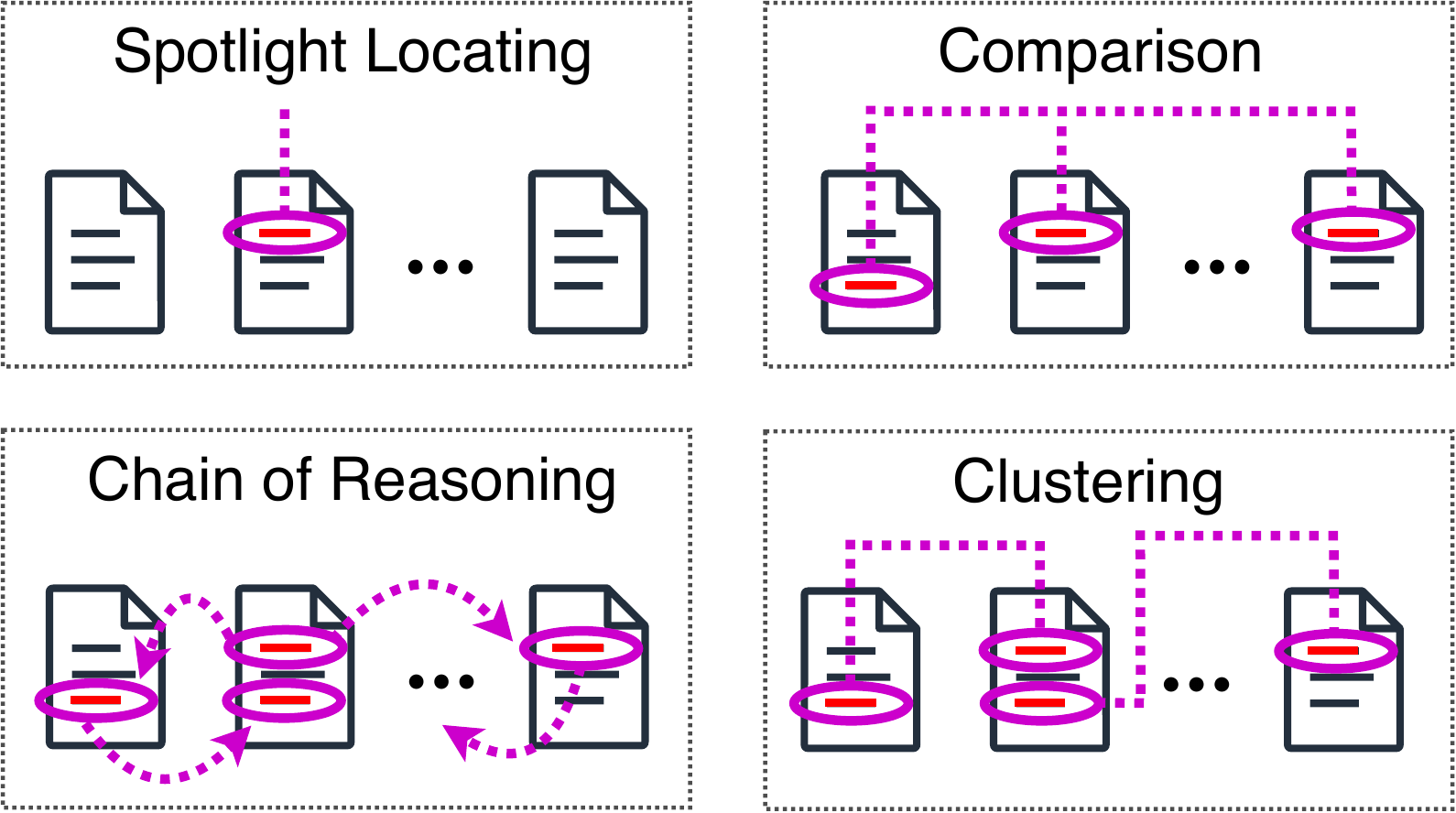}
\vspace{-18pt}
\captionof{figure}{Reasoning facets for each example in Loong~\citep{DBLP:conf/emnlp/WangCCL0WYXZLLY24}.}
\label{fig:reasoning_tasks}
\end{minipage}
\end{figure*}

\vspace{-1.633mm}
\subsection{Can't Claude Code Do It?}\label{sec:cant-claude-code}

% With the remarkable progress in coding agents~\citep{clark2026trillion}, a natural question is whether work like ours is even necessary at all; perhaps the simplest solution is enough: place all documents in a folder, provide the task, and \q{ask Claude Code}. We therefore test exactly this setup on CorpusQA, using the full Claude Code agentic harness (without web access tools; see \S\ref{appendix:claude-code-setup} for more details) using \texttt{claude-sonnet-4.6}~\citeyearpar{ClaudeSonnet46} and the latest \texttt{claude-sonnet-5}~\citeyearpar{ClaudeSonnet5}.

\vspace{-1.633mm}
With the remarkable progress in coding agents~\citep{clark2026trillion}, a natural question is whether a simpler approach might suffice: place all documents in a folder, provide the task, and \q{ask Claude Code}. We therefore test exactly this setup on CorpusQA with Claude Code (see \S\ref{appendix:claude-code-setup} for details).

\noindent\textbf{\method{} with a 3B-active model surpasses Claude Code with frontier models in accuracy.}
\method{} achieves 71\% accuracy using \texttt{Qwen3.5-35B-A3B} (3B active parameters). In comparison, Claude Code achieves 67\% with \texttt{Claude-Sonnet-4.6} and 70\% with \texttt{Claude-Sonnet-5} (Fig.~\ref{fig:claude-code}).

% \method{} outperforms Claude Code at $3.8\times$ lower cost.

\noindent\textbf{\method{} costs $3.7\times$ to $3.9\times$ less than Claude Code.}
The improvements come at lower cost: \method{} costs \$439, compared with \$1,625 for Claude Code with \texttt{Sonnet-5} and \$1,714 with \texttt{Sonnet-4.6}, making \method{} $3.7\times$ and $3.9\times$ cheaper, respectively, while achieving higher accuracy (Fig.~\ref{fig:claude-code}).

\vspace{-0.933mm}
\subsection{Ablation Studies}\label{sec:ablation}

\vspace{-0.333mm}
We ablate the two stages of \method{} which construct the representation used for reasoning (Fig.~\ref{fig:r3con}). 

\begin{wraptable}[7]{r}{0.45\textwidth}
\centering
\vspace{-7.5mm}
\caption{Accuracy in the ablation of the two phases of \method (see Fig.~\ref{fig:r3con}).}
\vspace{3mm}
\label{tab:ablation-draft}
\resizebox{\linewidth}{!}{%
\begin{tabular}{lcc}
\textbf{Variant}
& \textbf{Loong ($\uparrow$)}
& \textbf{CorpusQA ($\uparrow$)} \\
\midrule
\method{} (default) & 37.43\% & 70.64\% \\
\noalign{\vskip 2pt}
\arrayrulecolor{gray!150}\cdashline{1-3}[0.6pt/2pt]\arrayrulecolor{black}
\noalign{\vskip 3pt}
\hspace{1em}w/o extracting relevance & 23.96\%\loss{13.47} & 52.58\%\loss{18.06} \\
\hspace{1em}w/o structuring & 32.97\%\loss{4.46} & 63.22\%\loss{7.42} \\
\bottomrule
\end{tabular}%
}
\end{wraptable}

\noindent\textbf{Removal of extracting relevance or structuring degrades performance.} As shown in Tab.~\ref{tab:ablation-draft}, removing extracting relevance has the largest effect. Notably, without this stage, \method{} becomes similar to StructRAG~\citep{DBLP:conf/iclr/LiC0L0T0H0L25}, and indeed achieves comparable performance (Tab.~\ref{tab:overall-results}). Removing structuring also consistently degrades performance.

\vspace{-0.633mm}
\subsection{Understanding Why \method{} Succeeds While Others Fail}\label{sec:qualitative-analysis}

\vspace{-0.333mm}
In this section, we return to our running example (Fig.~\ref{fig:dracula-question}) and investigate why the nine baselines considered in this paper fail to attribute all deaths to Dracula, whereas \method{} succeeds.

\begin{wraptable}[15]{r}{0.4\textwidth}
\vspace{-8.2mm}
\caption{Taxonomy of baselines according to the principles (\S\ref{sec:computational-principles}) satisfied by their \textbf{representation-construction} phase. See \S\ref{appendix:baselines} for detailed analysis. \pmark{} = limited support; N/A = not applicable.}
\vspace{3mm}
\label{tab:method-properties}
\centering
\resizebox{\linewidth}{!}{%
\begin{tabular}{lccc}
\toprule
\textbf{Method}
& \makecell{\textbf{Task-} \\ \textbf{specific}}
& \makecell{\textbf{Enactive}}
& \makecell{\textbf{Flexible} \\ \textbf{Structuring}} \\
\midrule
RAPTOR~\citeyearpar{DBLP:conf/iclr/SarthiATKGM24} & \xmark & \cmark & \xmark \\
ReadAgent~\citeyearpar{DBLP:conf/icml/LeeCFCF24} & \xmark & \xmark & \xmark \\
MemAgent~\citeyearpar{yu2026memagent} & \cmark & \pmark & \xmark \\
HippoRAG2~\citeyearpar{DBLP:conf/icml/GutierrezSQZ025} & \xmark & \pmark & \xmark \\
LinearRAG~\citeyearpar{zhuang2026linearrag} & \xmark & \pmark & \xmark \\
StructRAG~\citeyearpar{DBLP:conf/iclr/LiC0L0T0H0L25} & \cmark & \xmark & \cmark \\
CodeAgent~\citeyearpar{smolagents} & N/A & N/A & N/A \\
RLM~\citeyearpar{DBLP:journals/corr/abs-2512-24601} & N/A & N/A & N/A \\
Claude Code~\citeyearpar{ClaudeCode} & N/A & N/A & N/A \\
A-RAG~\citeyearpar{DBLP:journals/corr/abs-2602-03442} & \xmark & \xmark & \xmark \\
\method{} (ours) & \cmark & \cmark & \cmark \\
\bottomrule
\end{tabular}%
}
\end{wraptable}

\noindent\textbf{Only \method{} satisfies all design principles.} Recall from \S\ref{sec:computational-principles} that \circled{1} extracting relevance requires being both \textit{task-specific} and \textit{enactive} and \circled{2} structuring should be \textit{flexible}. Our method directly instantiates each of these principles through its two stages, whose interplay allows \method{} to answer this task correctly, as illustrated throughout \S\ref{sec:method}. We further analyze each baseline in \S\ref{appendix:baselines} and find that none satisfy all principles, as shown in Tab.~\ref{tab:method-properties}.

\noindent\raisebox{-0.28em}{\includegraphics[height=1.50em]{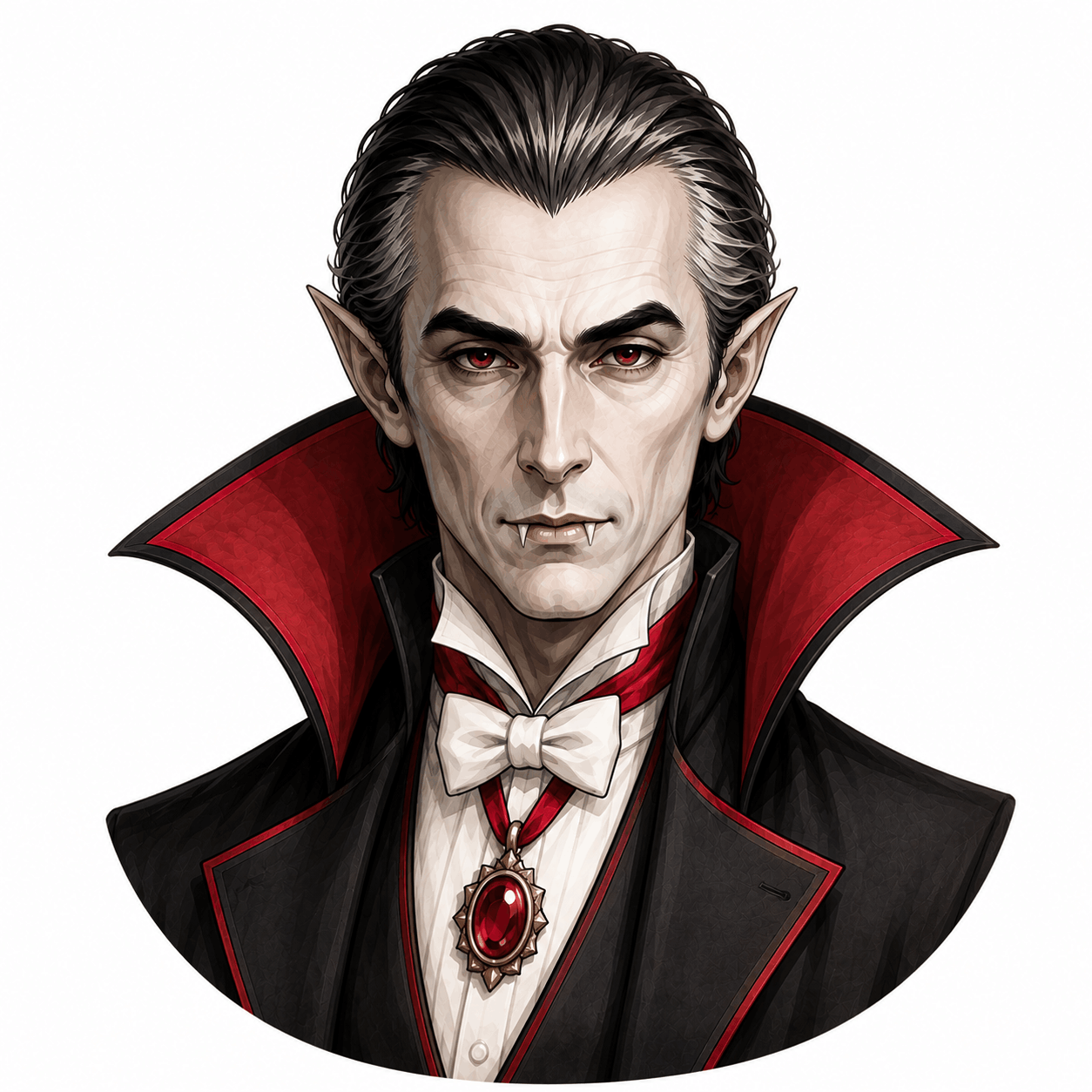}}\hspace{0.25em}\textbf{Why did Dracula get away with it?} We focus on several representative baselines to illustrate how their failures relate to missing or limited support for particular principles. First, RAPTOR~\citeyearpar{DBLP:conf/iclr/SarthiATKGM24} creates its representations (i.e., hierarchical summaries) \textbf{task-agnostically}. With no signal for what is relevant to the task, it discards details such as the specific crewmen who died aboard the Demeter and how many initially set sail (Fig.~\ref{fig:raptor-demeter-layer-one}). Once discarded, the solution becomes inaccessible.

StructRAG~\citeyearpar{DBLP:conf/iclr/LiC0L0T0H0L25} is task-specific, but has \textbf{no enactive mechanism} for reinterpreting earlier documents in light of new information. It discards crucial details about the captain's plan to tie his \q{hands to the wheel} until his \q{strength begins to fail}, as these only become relevant later when a newspaper reports a \q{dead helmsman} at the wheel. StructRAG therefore misses the captain's death (Fig.~\ref{fig:structrag-demeter-tables}). So do other graph-based approaches (e.g., HippoRAG2 does not even link Dracula to the Demeter events at all). A better, but still \textbf{limited},~instantiation of \textbf{enactivity} comes from MemAgent~\citeyearpar{yu2026memagent}, which maintains an evolving textual memory as it reads  through the corpus. When it reads the ship's log, it does not yet know enough to attribute these deaths to Dracula, and information about the deaths gradually fades from its memory until it is forgotten altogether (Fig.~\ref{fig:memagent-failure}).

Lastly, coding agents have \textbf{no constructed representation} at all, and instead rely on the model to dynamically discover an effective reasoning strategy. These agents all miss various multi-hop connections required by the task (\S\ref{appendix:evidence_attribution}). For example, the captain's death requires a two-hop connection, which even the best-performing coding agent, RLM~\citeyearpar{DBLP:journals/corr/abs-2512-24601}, fails to make (Fig.~\ref{fig:rlm-demeter-turns}).

\vspace{-0.933mm}
\section{Related Work}\label{sec:background}

\vspace{-0.633mm}
Work on reasoning over large contexts has largely followed two complementary directions: explicit representation construction to aid downstream reasoning and dynamic agentic reasoning.

\noindent\textbf{Explicit Representation-Construction.} In these approaches, the strategy for constructing a representation of the context is explicitly encoded in the system design (i.e., the harness). Some approaches construct and update textual memories as the context is processed~\citep{DBLP:conf/nips/XuLMGTZ25, hu2026memoryageaiagents, yu2026memagent}, with other work constructing offline summaries at different levels of granularity~\citep{DBLP:conf/iclr/SarthiATKGM24, DBLP:conf/icml/LeeCFCF24}. Other approaches choose to impose more explicit structure: GraphRAG and follow-up work use graphs~\citep{DBLP:journals/corr/abs-2404-16130, 3737916.3739818, DBLP:conf/icml/GutierrezSQZ025, zhuang2026linearrag}, alongside broader efforts to improve RAG systems~\citep{DBLP:conf/emnlp/LiDJZZZZD25, DBLP:conf/emnlp/HuangHYPCMCC25, wu2026findingmattersanchoringcontext, jin-etal-2026-sara, zhou-etal-2026-aed}. More recently, some methods make the structuring decision itself dynamic, including StructRAG~\citep{DBLP:conf/iclr/LiC0L0T0H0L25}, Structure-R1~\citep{DBLP:journals/corr/abs-2510-15191}, and LogicRAG~\citep{DBLP:conf/aaai/ChenZYZCCXCH26}. Taken together, these approaches provide effective but largely ad hoc answers to how context should be represented. Our work instead proposes principles that cut across these individual approaches, providing general guidance for how these representation-construction decisions should be made and why.

\noindent\textbf{Agentic Reasoning.}  In contrast, coding and tool-calling agents~\citep{DBLP:conf/iclr/YaoZYDSN023, DBLP:conf/icml/WangCY0L0J24, ClaudeCode, DBLP:journals/corr/abs-2512-24601, DBLP:journals/corr/abs-2605-05242, DBLP:journals/corr/abs-2602-03442} give models full freedom to structure their reasoning strategy at test time. Yet, this flexibility is a double-edged sword: the model must discover effective reasoning strategies on its own and there is increasing evidence that models struggle to do so in knowledge-intensive tasks~(\S\ref{sec:cant-claude-code};~\citealp{DBLP:journals/corr/abs-2511-16660,light2026deepreasoninggeneralpurpose, DBLP:conf/acl/TheologitisDSS26, xu-etal-2026-tps}). In our work, \method{} uses a coding agent only \textit{after} an effective representation has been constructed, relieving the model of having to discover one on its own inside the coding loop. More broadly, our principles could also be used to create training trajectories that teach models how to organize context effectively at test time.

% In this way, we relieve the model of having to discover an effective representation, and instead direct its \textit{dynamic} reasoning toward exploring it instead.

% In this way, we releive the model of ha

\vspace{-0.533mm}
\section{Conclusions}\label{sec:discussion}

\vspace{-0.233mm}
In this paper, we study the problem of organizing large heterogeneous contexts into representations that enable accurate downstream reasoning. Grounded in the cognitive theory of relevance realization, we propose a set of design principles to build rich task-specific and structured context representations. We then introduce \method{} which operationalizes them, and evaluate it against nine baselines on hard long-context  benchmarks, outperforming prior state-of-the-art methods by up to~$20$ percentage points. 

By keeping representation construction explicit, \method{} provides two important benefits. First, it keeps decisions about what is relevant and how information should be organized external to the model, where they can be inspected, adapted, and controlled by the data owner, rather than relying entirely on opaque relevance patterns internalized by the model during training. Second, better context representation reduces  reliance on model scale: we demonstrate how \method{} enables models an order of magnitude smaller to outperform frontier systems using much larger models (\S\ref{sec:overall-performance-and-cost} and \S\ref{sec:cant-claude-code}). This could enable data owners to deploy smaller models on their own infrastructure, retaining control over sensitive knowledge while still achieving frontier-level performance. These benefits are especially valuable in domains like science, medicine, law, finance, and journalism, which often depend on sensitive knowledge that must be handled with care. Explicit representation construction and smaller, locally deployable models could provide a path toward AI systems that are not only capable, but also more customizable and amenable to inspection and control.

\subsection*{AI use statement}

In this work, we used generative AI tools to improve the readability and wording of the manuscript. All AI-assisted edits were reviewed by the authors for correctness and consistency with the intended meaning. We take responsibility for the final content of this work.

% In this work, we did not use generative AI tools for tasks requiring disclosure. Additionally, we used generative AI tools to polish the writing of the manuscript. We have reviewed all AI-assisted work. AI-assisted edits to the manuscript were reviewed by the authors for correctness and consistency with the intended meaning. We take responsibility for the final content of this work, including text, claims or artifacts produced with the aid of generative AI.

\subsection*{Reproducibility statement}

We have made an effort to make this work as reproducible as possible. All prompts used by our method are provided in \S\ref{appendix:prompts}. Implementation details for all baseline methods are documented in \S\ref{appendix:baselines}. Benchmark details and additional evaluation metrics are provided in \S\ref{appendix:benchmarks}. Finally, all evaluation code is available at \url{\evalurl}.

% \subsection*{Acknowledgements}

% This research was supported by NSF III 2507117 and NSF IIS 2314527. This research was developed with funding from the Defense Advanced Research Projects Agency's (DARPA) SciFy program (Agreement No. HR00112520300). This material is based upon work supported in part by the Defense Advanced Research Projects Agency and the Air Force Research Laboratory, contract number(s): FA8650-23-C-7316. The views expressed are those of the author and do not reflect the official policy or position of the Department of Defense or the U.S.~Government.

\FloatBarrier

\bibliography{references}
\bibliographystyle{iclr2027_conference}

\FloatBarrier

\appendix

\startcontents[appendices]

\section*{Appendix Contents}

\printcontents[appendices]{}{1}{}

\section{Limitations}

We discuss several limitations and open questions that also motivate directions for future work.

\noindent \textbf{Reusing representations across similar tasks.} We build each representation based on the task, following our first principle of being task-specific. This means that \method{} needs to recompute the representation (i.e., relevant context and structure) for each new query. In terms of efficiency, this is more expensive than methods that can construct a generic, task-agnostic representation and reuse it across different tasks; still, such an approach risks creating a subpar representation which can hurt downstream reasoning, as we see in \S\ref{sec:results}. In future work, \method{} could embed and store relevance snippets and structures across queries, retrieving and reusing them for similar questions rather than recomputing everything from scratch. Over time, this could result in a rich collection of task-specific representations that could be combined and explored to provide an even better view of the context.

\noindent \textbf{Budgeting attention and intelligence in relevance extraction.} In every iteration of relevance realization in \method{}, the relevance snippets from all documents are fed back into the model. This assumes that the model's context window is large enough to accommodate them. In Appendix~\ref{appendix:size-repr}, we show that, in practice, the combined size of all relevance snippets across all documents remains roughly $1\%$ of the original input size for each example, giving substantial headroom as context windows continue to grow. However, scaling to unbounded corpora such as the open web will eventually require mechanisms that allocate attention more selectively. One avenue is more sophisticated enactive mechanisms that use the understanding developed in each round to discard documents deemed irrelevant, reducing the context that subsequent rounds need to process.

\section{Running Example}\label{appendix:running-example}

\subsection{Bram Stoker's Dracula (1897)}

Dracula~\citeyearpar{stoker1897dracula} is written in epistolary style---the story is assembled from a collection of documents (e.g., journals, telegrams, letters, etc.) produced by its characters as events unfold. The story is told in a roughly chronological order by piecing together relevant snippets from the different text artifacts. For example, the events of September 17--18 are told by piecing together entries from Lucy Westenra's and Dr.\ Seward's diaries, a telegram from Van Helsing to Seward, a memorandum left by Lucy describing the events of the night, and a letter from Mina to Lucy.

No individual document contains the complete story. Each records what its author knew at a particular point in time, often without understanding the significance of what they observed and with incomplete information. For example, when Mina discovers two small wounds on Lucy's neck, she attributes them to accidentally pricking Lucy with a safety-pin. Only when this observation is considered alongside later accounts of Lucy's deteriorating health and Dracula's activities does its significance become clear. Similarly, Dracula's arrival in England is recorded independently through the ship's log, a newspaper account of the shipwreck, and Mina's observations in Whitby. Understanding what happened therefore requires connecting information written by different people, at different times, and for different purposes.

\vspace{1.5mm}
\indent \textbf{Documents.} For our purposes, we reconstruct the individual text artifacts that make up the novel. For example, Jonathan Harker's journal appears throughout the book as a series of dated entries. We collect and concatenate these entries to reconstruct Jonathan Harker's journal---a single document. We do the same for the other recurring journals and diaries. Standalone artifacts, such as individual letters, telegrams, memoranda, and newspaper cuttings, are treated as separate documents. We end up with a collection of 46 documents of different types, summarized in Tab.~\ref{tab:dracula_documents}. 

\begin{table}[t]
\centering
\small
\begin{tabular}{lcl}
\toprule
\textbf{Document type} & \textbf{\# Docs.} & \textbf{Description} \\
\midrule
Journal / diary    & 5  & Personal journals and diaries (e.g., Jonathan Harker's journal) \\
Letter             & 20 & Letters between characters (e.g., Mina to Lucy) \\
Telegram           & 8  & Telegrams between characters (e.g., Van Helsing to Seward) \\
Memorandum         & 4  & Written accounts of events (e.g., Lucy's final memorandum) \\
Newspapers  & 4  & Newspaper cuttings (e.g., The Pall Mall Gazette) \\
Note               & 2  & Short written notes (e.g., Van Helsing's contingency note) \\
Ship's log         & 1  & The log of the Demeter ship (from the captain) \\
Report             & 1  & Medical report (Dr.\ Hennessey to Dr.\ Seward) \\
Phonograph & 1  & Transcribed recording of Van Helsing \\
\midrule
\textbf{Total \#Docs}     & \textbf{46} & \\
\textbf{Total Tokens}     & \textbf{220k} & \\
\bottomrule
\end{tabular}
\caption{The reconstructed document collection in Bram Stoker's Dracula~\citeyearpar{stoker1897dracula}.}
\label{tab:dracula_documents}
\end{table}

\vspace{1.5mm}
\indent \textbf{Task.} We consider the following query over the resulting corpus:
\begin{quote}
\textit{On the voyage to England and during his stay there, how many people did Dracula kill, either directly or indirectly, and who were they?}
\end{quote}
\vspace{1.5mm}
\indent \textbf{Answer.} The answer is 13: the 9 members of the \textit{Demeter}'s crew, Mr.\ Swales, Mrs.\ Westenra, Lucy Westenra, and Renfield.

Of course, this answer does not appear explicitly in any individual document. In fact, each individual death follows a lot of highly non-trivial reasoning steps throughout the different documents (\S\ref{appendix:evidence_attribution}).

\subsection{Evidence Attribution}\label{appendix:evidence_attribution}

We trace the relevant evidence for each of the 13 deaths below.

\vspace{1.5mm}
\noindent \textbf{Demeter's crew (9).} The main evidence is scattered throughout the ship's log. First, we must establish that the ship leaves Varna with nine people aboard: the captain says \q{On 6 July we finished taking in cargo, silver sand and boxes of earth.} and then \q{At noon set sail. East wind, fresh. Crew, five hands ... two mates, cook, and myself (captain).} Over the course of the voyage, the captain records the crew disappearing or dying one by one: \q{one of crew, Petrofsky, was missing} (16 July); \q{another man lost--disappeared} (24 July); \q{Another tragedy ... Are now without second mate} (29 July); \q{both man of watch and steersman missing. Only self and mate and two hands left to work ship} (30 July); \q{One more gone. Lord, help us!} (2 August); and \q{I went to relieve the man at the wheel, and when I got to it found no one there} (3 August). Later that same day, the mate \q{sprang on the bulwark and deliberately threw himself into the sea}. By the end, only the captain remains.

We must also establish that Dracula is aboard the ship. The crew repeatedly report seeing a mysterious man, including \q{a tall, thin man, who was not like any of the crew, ..., go along the deck forward, and disappear.}, \q{I saw It, like a man, tall and thin, and ghastly pale.}, and, finally, \q{in the dimness of the night I saw It--Him!} Other accounts identify this figure as Dracula: Seward later encounters a \q{tall, thin man, clad in black} whom he and the others recognize as the Count.

The captain is the last remaining member of the crew. His final log entry states: \q{I shall tie my hands to the wheel when my strength begins to fail.} His death, however, can only be confirmed by a different document: an English newspaper cutting from The Dailygraph describes the ship's arrival with \q{lashed to the helm was a corpse, with drooping head, which swung horribly to and fro at each motion of the ship.} Together, the log and newspaper account identify the otherwise unnamed corpse as the captain.

\vspace{1.5mm}
\noindent \textbf{Mr.\ Swales (1).} The Demeter brings Dracula from Transylvania to England, arriving in the coastal town of Whitby. While staying there, Mina befriends Mr.\ Swales, an elderly local whom she frequently meets at their favorite seat in the churchyard overlooking the harbor. Shortly after the ship's arrival, Mina records that \q{poor old Mr.\ Swales was found dead this morning on our seat, his neck being broken.} The circumstances of his death are unusual: according to the doctor, he had \q{fallen back in the seat in some sort of fright,} with a look of \q{fear and horror on his face.} Mina herself wonders, \q{Perhaps he had seen Death with his dying eyes!}

Connecting his death to Dracula requires evidence surrounding the location and timing. When the Demeter arrives two days earlier, a newspaper account describes an \q{immense dog} leaping from the ship and \q{making straight for the steep cliff, where the churchyard hangs over the laneway to the East Pier.} The day after Swales is found dead, Mina returns to the same seat and finds Lucy there asleep. She observes \q{something dark} bending over Lucy and, as she approaches, sees \q{a white face and red, gleaming eyes.} It can be safely assumed that the mysterious dog that came from the ship was Dracula himself in one of his guises, and that it was responsible for knocking the old man down and causing his death.\footnote{\url{https://www.cliffsnotes.com/literature/d/dracula/summary-and-analysis/chapters-78}}

\vspace{1.5mm}
\noindent \textbf{Mrs.\ Westenra (1).} Mrs.\ Westenra is Lucy's mother, with whom Lucy lives at Hillingham. She suffers from a serious heart condition. And, Mina Murray notes (in her journal) that \q{her heart is weakening} and that \q{a sudden shock would be almost sure to kill her.} On the night of September 17, while Mrs.\ Westenra is in Lucy's room, a wolf crashes through the window. In a memorandum left by Lucy, she records that \q{in the aperture of the broken panes there was the head of a great, gaunt grey wolf.} Her mother \q{cried out in a fright} and then suddenly collapsed; Lucy realizes shortly afterward that \q{her dear heart had ceased to beat.}

Connecting the wolf to Dracula requires a separate newspaper account. In an interview with the zookeeper published in The Pall Mall Gazette, a man matching Dracula's appearance visits the wolf enclosure earlier that day. The keeper describes him as \q{a tall, thin chap, with a 'ook nose and a pointed beard} with \q{red eyes} and \q{a mouth full of white, sharp teeth.} The same newspaper account reports that a wolf named Bersicker escapes from the enclosure that night. After Mrs.\ Westenra's death, Bersicker returns to the zoo with his head \q{all cut and full of broken glass}---matching Lucy's memorandum describing the wolf crashing through her window. Together, these accounts connect Dracula to the escaped wolf whose attack frightens Mrs.\ Westenra and kills her indirectly.

\vspace{1.5mm}
\noindent \textbf{Lucy Westenra (1).} Lucy is Mina's close friend and becomes Dracula's primary victim after his arrival in England. The first attack occurs in the Whitby churchyard. Mina's journal describes finding Lucy asleep on their favorite seat with \q{something dark} bending over her and later notices \q{two little red points like pin-pricks} on Lucy's neck. Mina initially assumes that she caused them herself with a safety-pin, but later observes that \q{the tiny wounds seem not to have healed. They are still open, and, if anything, larger than before.}

Over the following weeks, Lucy becomes increasingly weak and continues to lose blood with no apparent explanation. In her own diary, she records hearing \q{scratching or flapping at the window} during the night and waking \q{horribly weak.} Dr.\ Seward's diary records that her condition becomes so severe that she receives blood transfusions from four different men, yet the blood repeatedly seems to disappear. Together with the persistent wounds on her neck and the nighttime encounters, these accounts establish that Dracula continues to feed on Lucy. Her condition eventually deteriorates, and on September 20, Dr.\ Seward's diary records her death: \q{And then Lucy's breathing became stertorous again, and all at once it ceased. It is all over,' said Van Helsing. She is dead!'}

\vspace{1.5mm}
\noindent \textbf{Renfield (1).} Renfield is a patient in Dr.\ Seward's asylum who initially regards Dracula as his \q{Master}, but eventually turns against him after realizing that Dracula has been attacking Mina. While mortally wounded, Renfield recounts their confrontation to Seward: \q{it made me mad to know that He had been taking the life out of her} and, when he tried to stop Dracula, \q{He raised me up and flung me down.} Seward's diary records that Renfield's \q{back is broken} and that \q{the real injury was a depressed fracture of the skull.} His death is later confirmed in the same diary when Arthur reports that \q{the poor fellow is dead.}

\vspace{1.5mm}
\noindent \textbf{Excluded Deaths.} Identifying the victims also requires reasoning about which deaths should \textit{not} be counted. Four deaths connected to Dracula fall outside the question's time window. Before the Demeter sets sail, at the castle in
Transylvania, Dracula brings an infant to the three vampire sisters, followed by the child's mother, who comes to the castle crying \q{Monster, give me my child!} and is killed by the wolves he summons. After Dracula flees England, Petrof Skinsky, the agent hired to move his last box, is murdered---a death Mina explicitly attributes to Dracula, who \q{blotted out his traces, as he thought, by murdering his agent}. Quincey Morris also dies after Dracula's flight, and is killed by the Szgany guarding Dracula's cart rather than by Dracula himself. Two other deaths also happen within the time window but must still be excluded. Mr.\ Hawkins \q{has died very suddenly} in Exeter, with no evidence connecting his death to Dracula, while Arthur's father dies after a longer decline documented across several accounts. Finally, several apparent victims do not die at all: the Hampstead children survive their attacks, as does Mina, who remains alive in the closing note seven years later.

\subsection{Scoring}

We score answers using \texttt{gpt-5.4-mini} as an LLM judge. We count an answer as correct only if it identifies both the correct number of victims and their individual names: the nine victims aboard the Demeter, Mrs.\ Westenra, Mr.\ Swales, Lucy, and Renfield. Because Mr.\ Swales' death is more inferential and could be contested, we also evaluated scores when excluding him from the ground truth and found that this does not change the scores of either \method{} or any baseline.

\subsection{Parametric Knowledge}

Naturally, we expect \textit{Dracula}~\citep{stoker1897dracula} to appear in the models' pre-training data and thus be part of their parametric knowledge. We therefore test whether models can answer the Dracula question (Fig.~\ref{fig:dracula-question}) under two settings: \circled{1} using parametric knowledge alone, where we ask the same question while specifying \q{In Bram Stoker's 1897 novel Dracula, [...]} to distinguish the novel from other versions; and \circled{2} with the entire novel provided in context. We conduct these experiments using \texttt{Qwen3.5-35B-A3B}~\citep{qwen3.5}, the default model used throughout our experiments.

\vspace{1.5mm}
\noindent \textbf{Results.} We find that \texttt{Qwen3.5-35B-A3B} fails to answer the question correctly in both settings: when relying on parametric knowledge alone and when provided with the entire novel in context (Tab.~\ref{tab:parametric-knowledge}; see also the raw model responses in Fig.~\ref{fig:parametric-knowledge-qwen-outputs}).

\begin{table}[t]
\centering
\small
\caption{Answers to the Dracula question using parametric knowledge alone and with the full novel provided in context. The ground-truth answer is 13, but also requires the correct people named. Full model responses are shown in Fig.~\ref{fig:parametric-knowledge-qwen-outputs}.}
\vspace{3mm}
\label{tab:parametric-knowledge}
\begin{tabular}{lcccc}
\toprule
\multirow{2}{*}{\textbf{Model}}
& \multicolumn{2}{c}{\textbf{Parametric knowledge}}
& \multicolumn{2}{c}{\textbf{With corpus in context}} \\
\cmidrule(lr){2-3}\cmidrule(lr){4-5}
& \textbf{Answer} & \textbf{Correct}
& \textbf{Answer} & \textbf{Correct} \\
\midrule
\texttt{Qwen3.5-35B-A3B}~\citeyearpar{qwen3.5} & 31 & \xmark & 7 & \xmark \\
\bottomrule
\end{tabular}
\end{table}

\begin{figure*}[t]
\centering
\begin{minipage}[t]{0.485\textwidth}
\centering
\textbf{Parametric knowledge}
\vspace{1mm}
\begin{minted}[fontsize=\tiny, frame=single, breaklines=true, breaksymbol={}, escapeinside=||]{text}
Based on Bram Stoker's 1897 novel *Dracula*, the count of people killed by the vampire, either directly or indirectly, can be broken down into three specific groups: the crew of the *Demeter*, Lucy Westenra, and Dr. John Seward's patient, Mark Renfield.

While exact numbers regarding the crew are slightly ambiguous in the text, the most consistent count provided within the narrative is **|\textcolor{red}{\bfseries 31}| people**.

Here is the detailed breakdown:

### 1. The Crew of the *Demeter* (29 Men)
The novel begins with the arrival of the ship *Demeter* at Whitby. Throughout the voyage, Dracula (traveling in the form of a dog and later a man inside a crate) slaughtered the crew to feed.
*   **Location:** On the voyage from Varna to Whitby, and the ship itself.
*   **Cause of Death:** All crew members were found dead upon the ship's arrival. The Captain's log entries describe the men disappearing one by one, and the final log states, "We are all dead."
*   **Number:** According to Renfield's journal in Chapter 22, there were **"29 men on board"** the *Demeter* (a figure generally accepted to include the Captain and his crew). The Whitby report in Chapter 8 confirms that the entire crew, including the captain, was found dead.
*   **Status:** **Direct Kills** (Slaughtered by Dracula).

### 2. Lucy Westenra (1 Woman)
*   **Location:** Whitby and London.
*   **Cause of Death:** Lucy was bitten by Dracula on multiple occasions (directly), which turned her into a vampire. While she was physically "killed" by being staked through the heart by Arthur Holmwood and the other vampire hunters, her death was the direct result of Dracula's transformation of her.
*   **Status:** **Indirect Kill** (Dracula's bite caused the condition that led to her death; she was killed to prevent her from harming others).

### 3. Mark Renfield (1 Man)
*   **Location:** London (The Crippled Asylum / Seward's Asylum).
*   **Cause of Death:** Renfield was Dracula's primary human contact in London. While he did not die from a physical attack by Dracula's fangs in the asylum, Van Helsing and the narrative confirm that Dracula was psychically "feeding" on Renfield. This psychic draining drained Renfield's vitality, leading to his death in his asylum bed.
*   **Status:** **Indirect Kill** (Psychic draining and influence).

### Total Count
**29 (Crew) + 1 (Lucy) + 1 (Renfield) = 31 people.**

**Important Notes on Exclusions:**
*   **Mina Harker:** Although Dracula bit Mina and infected her with blood, she was never turned and survived the novel. Therefore, she is not counted as a kill.
*   **Other Characters:** No other minor characters (such as Carfax servants or patients at the asylum) are confirmed to have died as a result of Dracula's actions in the text.

**Final Answer:**
In Bram Stoker's novel, Dracula is responsible for the deaths of **|\textcolor{red}{\bfseries 31 people}|** directly or indirectly. These are:
1.  **29 crew members** of the *Demeter* (slaughtered during the voyage).
2.  **Lucy Westenra** (turned into a vampire by Dracula and subsequently staked).
3.  **Mark Renfield** (died due to psychic draining by Dracula).
\end{minted}
\end{minipage}
\hfill
\begin{minipage}[t]{0.485\textwidth}
\centering
\textbf{With corpus in-context}
\vspace{1mm}
\begin{minted}[fontsize=\tiny, frame=single, breaklines=true, breaksymbol={}, escapeinside=||]{text}
Based on the provided documents, Dracula killed or caused the death of the following people:

**On the Voyage to England (Ship** ***Demeter*):**
*   **Seven (7) Crew Members:** The Log of the "Demeter" (Document 6) details the disappearance and death of the crew. Specifically mentioned or implied victims include:
    *   **Petrofsky** (Hand, missing 16 July).
    *   **One man** (missing 24 July).
    *   **One man** (missing 29 July).
    *   **One man of watch and the Steersman** (missing 30 July).
    *   **The Second Mate** (died/committed suicide 2 August).
    *   **The Captain** (died on board 4 August).
    *   *(Total Crew Start: 9 (5 hands, 2 mates, 1 cook, 1 captain); Total Crew Dead: 7)*.

**During Stay in England:**
*   **Two (2) Adults:**
    *   **Mrs. Westenra** (Lucy's mother, died of fright/heart condition induced by the attack).
    *   **Lucy Westenra** (died 17 September, became an Un-Dead).
*   **Several (Unknown Number) Children:**
    *   Multiple newspaper clippings (Documents 12, 25, 35) report children being lured away by the "bloofer lady" (the Un-Dead Lucy/Dracula) with tiny wounds in the throat.
    *   Document 12 specifically mentions "Another child injured" and a child missed from Hampstead Heath.
    *   Document 25 mentions "Several cases" and "two children missed last night."
*   **Renfield (Highly Implied):**
    *   Dr. Seward's Diary (Document 14) describes Renfield found lying in a "glittering pool of blood" with a broken back and bruises, implying death by Dracula's attack (though the text focuses on the injury).

**Total Confirmed Dead:** |\textcolor{red}{\bfseries 9 Humans}| (7 on *Demeter*, 2 Westenras).
**Total Injured/Implied Dead:** Several Children, Renfield.
\end{minted}
\end{minipage}
\caption{Responses of our default model, \texttt{Qwen3.5-35B-A3B}, to the Dracula question (Fig.~\ref{fig:dracula-question}) using parametric knowledge alone (left) and with the full novel provided in context (right). Both responses are incorrect which showcases the difficulty of the Dracula task (see \S\ref{appendix:evidence_attribution}).}
\label{fig:parametric-knowledge-qwen-outputs}
\end{figure*}

\section{Experimental Details}

\subsection{Benchmarks}\label{appendix:benchmarks}

\subsubsection{CorpusQA} 

CorpusQA~\citep{lu2026corpusqa} is a corpus-level analysis benchmark built around questions whose answers require reading and aggregating information over an entire collection of documents. The benchmark is designed so that answer-critical evidence is highly dispersed and compositional. It covers financial, education, and real-estate domains. We use the one-million-token question split, as evaluating larger splits across all nine baselines exceeds our computational budget (e.g., the 4M split).

\vspace{2.201mm}
\noindent\textbf{Metrics.} We report accuracy and we follow the benchmark's llm-as-a-judge based scoring mechanism~\citep{lu2026corpusqa} using \texttt{gpt-5.4-mini} which decides whether the answer is correct or not (scoring each example 0 or 1).

\subsubsection{Loong} 

Loong~\citep{DBLP:conf/emnlp/WangCCL0WYXZLLY24} is a long-context multi-document QA benchmark spanning three professional domains: academic papers, financial reports, and legal cases. Questions fall into four reasoning categories: \emph{spotlight locating}, which requires finding evidence in a single document among similar decoys; \emph{comparison}, which requires locating and comparing values across documents; \emph{clustering}, which requires grouping or integrating evidence from several sources; and \emph{chain-of-reasoning}, which requires multi-hop or temporal reasoning across documents. These categories let us examine not only whether a method handles long context, but also which kind of dispersed reasoning it supports. The benchmark also varies context length by increasing the number and size of real documents in the bundle. This makes Loong useful for studying how performance changes as the amount of necessary context grows. We use the three out of four context sizes (50k-100k, 100k-200k, and 200k-250k tokens) and exclude the smallest (10-50k).

\vspace{2.201mm}
\noindent\textbf{Metrics.} We report accuracy using an LLM-as-a-judge and follow the original evaluation prompts~\citep{DBLP:conf/emnlp/WangCCL0WYXZLLY24}. The judge assigns a score to each answer, and we count only answers receiving a perfect score as correct as in \citep{DBLP:conf/iclr/LiC0L0T0H0L25}. We use \texttt{gpt-5.4-mini} as the judge.

\subsection{Baselines}\label{appendix:baselines}

\subsubsection{RAPTOR}

\vspace{2.201mm}
\noindent\textbf{Method.} To address tasks requiring a more holistic understanding of the corpus, \citet{DBLP:conf/iclr/SarthiATKGM24} propose RAPTOR, which recursively summarizes the corpus. It first segments the input into leaf chunks, embeds them, clusters semantically related nodes, and asks an LLM to summarize each cluster into a parent node. This cluster-and-summarize process repeats bottom-up, producing a tree in which leaves retain local detail and higher layers contain increasingly abstract summaries. At query time, RAPTOR collapses all tree nodes into a single retrieval pool and retrieves across every abstraction level before passing the selected nodes to a reader LLM.

\vspace{2.201mm}
\noindent\textbf{Principles.} RAPTOR constructs its textual representation (i.e., its tree of summaries) without knowledge of the downstream task, and is therefore not \emph{task-specific}. Its construction is \emph{enactive}: summaries are recursively constructed from and informed by the corpus as the hierarchy develops. Finally, its structure is fixed in advance to a hierarchy of textual summaries with a retrieval pool, and therefore is not \emph{flexible}. 

%Finally, reasoning is not \emph{dynamic}: at test time, RAPTOR performs a fixed top-$k$ retrieval over the constructed representation and answers from the retrieved nodes. \yell{check consistent with no reasoning + remove reasoning}

\vspace{2.201mm}
\noindent\textbf{Evaluation Details.} In our experimental evaluation, we replace the local SBERT encoder with the more powerful \texttt{text-embedding-3-small} from OpenAI. Furthermore, RAPTOR was proposed and evaluated on benchmarks far smaller than CorpusQA and Loong---for example, QuALITY~\citep{DBLP:conf/naacl/PangPJNPCPMT0B22}, with only a few thousand tokens per example. On our benchmarks, the default 100-token leaf size yields on the order of 10k leaves for a single example, requiring thousands of summary calls and making it impractical to run with large reasoning models. We therefore scale the leaf size to roughly 2,000 tokens, the cluster-split ceiling from roughly 3,500 to 40,000 tokens, the summary length from 100 to 400 tokens, and retrieval from the top 10 to the top 20 nodes. This configuration scales RAPTOR's hyperparameters to the corpus sizes of Loong and CorpusQA so that they are computationally feasible to run.

\subsubsection{ReadAgent}

\vspace{2.201mm}
\noindent\textbf{Method.} ReadAgent~\citep{DBLP:conf/icml/LeeCFCF24} constructs an offline gist memory that provides a compressed view of the corpus while selectively expanding back into the original documents. More specifically, it first divides the document(s) into semantically coherent pages using the model to choose where pages naturally end (pagination). Then, it summarizes each page into a short gist, and keeps the ordered list of gists as a single memory. At test time, the model reads this compact memory and selects a small number of pages to re-open in full which provide the context for answering.

\vspace{2.201mm}
\noindent\textbf{Principles.} ReadAgent constructs its textual representation (i.e., its gist memory) without knowledge of the downstream task, and is therefore not \emph{task-specific}. Its construction is not \emph{enactive}: each page is summarized independently, without further interaction with the corpus. Finally, its structure is fixed in advance to an ordered list of textual gists, and therefore is not \emph{flexible}. 

%Finally, its reasoning is limited in \emph{dynamicity}: the model can choose which pages to reopen, but does so only once and cannot continue opening pages as its reasoning unfolds. If it could iteratively open additional pages based on what it had learned, we would classify its reasoning as fully dynamic. \yell{check consistent with no reasoning + remove reasoning}

\vspace{2.201mm}
\noindent\textbf{Evaluation Details.} CorpusQA is much larger than the benchmarks on which ReadAgent was originally evaluated. In practice, we found that the original hyperparameters result in an impractical number of individual calls to the large reasoning model, so we scale them with the corpus size. Specifically, we scale the page size from roughly 600 to 6,000 words, the pagination thresholds from 280/350 to 2,800/3,500, and the per-gist target from roughly 128 to 640 tokens. This scaling allows ReadAgent to finish within a practical compute budget on CorpusQA. Loong is evaluated using the original hyperparameters.

\subsubsection{MemAgent}

\vspace{2.201mm}
\noindent\textbf{Method.} MemAgent~\citep{yu2026memagent} maintains a task-focused textual memory by processing documents sequentially and giving the model the ability to retain, discard, or condense information. More specifically, it streams the document collection through the model in fixed-size chunks while maintaining a short textual memory. For each chunk, the model receives the question, the current memory, and the current chunk, then rewrites the memory by retaining information it predicts will be useful and discarding the rest. After the final chunk, the answer is generated from the question and the final memory alone. 

\vspace{2.201mm}
\noindent\textbf{Principles.} MemAgent constructs its textual representation (i.e., its memory) with knowledge of the downstream task, and is therefore \emph{task-specific}. However, its construction is limited in \emph{enactivity}: although its memory evolves as it processes new chunks, it makes only a single linear pass through the corpus, without the ability to revisit or reinterpret earlier information as its understanding evolves. Finally, its structure is fixed in advance to a bounded textual memory, and therefore is not \emph{flexible}.

%Finally, reasoning is not \emph{dynamic}: after processing the corpus, MemAgent answers directly from its final memory with a single call to the LLM.

\vspace{2.201mm}
\noindent\textbf{Evaluation Details.} We use the released \texttt{RL-MemoryAgent-14B}\footnote{\url{https://huggingface.co/BytedTsinghua-SIA/RL-MemoryAgent-14B}} model, which is based on \texttt{Qwen2.5-14B} and trained with RL in their seminal work~\citep{yu2026memagent}.

\subsubsection{HippoRAG2}

\vspace{2.201mm}
\noindent\textbf{Method.} HippoRAG2~\citep{DBLP:conf/icml/GutierrezSQZ025} is a retrieval method inspired by associative human memory. It first constructs a knowledge graph from the corpus by extracting entity--relation triples from each passage and connecting semantically similar entities across passages. At test time, it identifies entities and passages relevant to the query and propagates their relevance through the graph using Personalized PageRank, allowing passages connected through intermediate concepts to surface. The top-ranked passages are finally read by an LLM to generate the answer.

\vspace{2.201mm}
\noindent\textbf{Principles.} HippoRAG2 constructs its graph representation without knowledge of the downstream task, and is therefore not \emph{task-specific}. Its construction has limited \emph{enactivity}: information extracted from each passage is progressively integrated with previously processed information through the shared graph. However, this interaction does not develop a broader contextual understanding that can be used to revisit and reinterpret earlier information as the corpus is processed. Finally, its structuring is not \emph{flexible}: HippoRAG2 pre-commits to organizing information as a knowledge graph.

% Finally, its reasoning has limited \emph{dynamicity}: the query influences online retrieval, including an LLM-based recognition step that filters relevant triples before graph search. However, these operations follow a fixed retrieval pipeline, rather than allowing the model to dynamically choose and revise its reasoning process based on intermediate results. \yell{check consistent with no reasoning + remove reasoning}

\vspace{2.201mm}
\noindent\textbf{Evaluation Details.} For embeddings, we use OpenAI's \texttt{text-embedding-3-small}. We segment CorpusQA and Loong into approximately 8,000- and 3,000--token chunks, respectively, chosen to keep the number of OpenIE calls computationally feasible for each benchmark.

\subsubsection{LinearRAG}

\vspace{2.201mm}
\noindent\textbf{Method.} Traditional retrieval-based methods can struggle when tasks require following chains of evidence distributed across many documents. \citet{zhuang2026linearrag} proposed LinearRAG, which first constructs a graph of entities from the corpus together with their relevant passages. At test time, it starts with entities mentioned in the query and progressively discovers relevant entities that extend the reasoning chain, extracting the passages needed to answer the query. Then, it ranks passages with semantic similarity to the query with the importance of the entities they contain (with some further refinements). The top-ranked passages are finally read by an LLM to generate the answer.

\vspace{2.201mm}
\noindent\textbf{Principles.} LinearRAG constructs its graph representation without knowledge of the downstream task, and is therefore not \emph{task-specific}. Its construction has limited \emph{enactivity}: information extracted from the corpus is integrated into a shared graph, but this accumulated information does not inform how the corpus is subsequently interpreted or represented. Finally, its structure is fixed in advance to a relation-free graph over entities, sentences, and passages, and therefore is not \emph{flexible}. 

%Finally, its reasoning has limited \emph{dynamicity}: LinearRAG progressively discovers relevant entities, with what it discovers informing where retrieval goes next. However, this adaptation takes place within a prescribed retrieval procedure. \yell{check consistent with no reasoning + remove reasoning}

\vspace{2.201mm}
\noindent\textbf{Evaluation Details.} In our evaluation, we use OpenAI's \texttt{text-embedding-3-small} in place of the paper's default sentence encoder.

\subsubsection{StructRAG}

\vspace{2.201mm}
\noindent\textbf{Method.} \mbox{StructRAG}\footnote{Despite its name, StructRAG is not a RAG-based method.}~\citep{DBLP:conf/iclr/LiC0L0T0H0L25} dynamically induces a task-specific structure for the corpus. Given the query and document contents, it first selects one of several representations---such as a table, graph, or catalog---suited to the task. It then transforms each document into the selected representation. Finally, a utilizer decomposes the query into sub-questions, extracts relevant information from the structured representations, and combines the results into a final answer.

\vspace{2.201mm}
\noindent\textbf{Principles.} StructRAG constructs its representation with knowledge of the downstream task, and is therefore \emph{task-specific}. Its construction is not \emph{enactive} since each document is transformed into the selected structure independently, without further interaction with the corpus. Finally, its structure is \emph{flexible}, as it dynamically chooses among several structures (e.g., table, graph, catalog) based on the task.

\vspace{2.201mm}
\noindent\textbf{Evaluation Details.} In our evaluation we increase the maximum generation length from 4,096 to 32,768 tokens to accommodate reasoning models.

% Finally, reasoning is not \emph{dynamic}: StructRAG follows a fixed procedure of decomposing the query into sub-questions, extracting relevant information from the constructed representation, and combining the results into an answer.

\subsubsection{CodeAgent}

\vspace{2.201mm}
\noindent\textbf{Method.} Coding agents~\citep{DBLP:conf/icml/WangCY0L0J24,smolagents} allow the model to interact symbolically with large corpora and dynamically structure its reasoning through executable code. 

\vspace{2.201mm}
\noindent\textbf{Principles.} CodeAgent does not construct an intermediate representation of the corpus.

\vspace{2.201mm}
\noindent\textbf{Evaluation Details.} We use the \texttt{CodeAgent} implementation from \texttt{smolagents}, a production-oriented implementation of the CodeAct paradigm~\citep{DBLP:conf/icml/WangCY0L0J24}. To handle long document bundles, we place the documents directly into the executor state.

On CorpusQA, we increase the maximum number of steps from the default of 20 to 60, as we found in initial experiments that the agent often required more than 20 steps. On Loong, we retain the default of 20 steps for the larger \texttt{Qwen3.5-35B-A3B}, which did not encounter this limitation. However, for the experiments with the smaller \texttt{Qwen3.5-4B} and \texttt{Qwen3.5-9B} models (see \S\ref{appendix:different-models}), we increase the limit to 60 steps on Loong as well.

\subsubsection{RLM}

\vspace{2.201mm}
\noindent\textbf{Method.}  
Recursive Language Models (RLMs)~\citep{DBLP:journals/corr/abs-2512-24601} allow an agent to recursively call itself within a Python environment. These recursive calls enable powerful task-decomposition patterns, including spawning sub-agents for individual sub-tasks. 

\vspace{2.201mm}
\noindent\textbf{Principles.} RLM do not construct an intermediate representation of the corpus.

\vspace{2.201mm}
\noindent\textbf{Evaluation Details.} We use the official \texttt{rlms} package from PyPI, version 0.1.2.

\subsubsection{A-RAG}

\vspace{2.201mm}
\noindent\textbf{Method.} The work of \citet{DBLP:journals/corr/abs-2602-03442} on A-RAG equips a ReAct-style~\citep{DBLP:conf/iclr/YaoZYDSN023} agent with three tools: keyword search, semantic search, and full-chunk reading. This allows the model to refine its search strategy as new evidence is discovered and connect multi-hop evidence chains as needed. 

\vspace{2.201mm}
\noindent\textbf{Principles.} A-RAG constructs its retrieval representation (i.e., passage embeddings and search indices) without knowledge of the downstream task, and is therefore not \emph{task-specific}. Its construction is not \emph{enactive}, as passages are indexed independently without further interaction with the corpus. Finally, its structure is fixed in advance to a retrieval index, and therefore is not \emph{flexible}.

\vspace{2.201mm}
\noindent\textbf{Evaluation Details.} In our evaluation, we replace the original \texttt{Qwen3-Embedding-0.6B} embedding model with OpenAI's \texttt{text-embedding-3-small}.

\subsubsection{Claude Code}\label{appendix:claude-code-setup}

\vspace{2.201mm}
\noindent\textbf{Method.} Claude Code~\citep{ClaudeCode} is Anthropic's coding agent harness, which allows the model to reason over files through a collection of coding and system tools. 

\vspace{2.201mm}
\noindent\textbf{Principles.} Claude Code does not construct an intermediate representation of the corpus.

\vspace{2.201mm}
\noindent\textbf{Evaluation Details.} For each task, we write the document bundle to a temporary working directory and run the Claude Code CLI in headless mode, instructing the agent to answer using only the provided files. The agent has full permissions within this directory and can interact with the corpus using its standard toolset.

We use Claude Code version 2.1.181 with its default toolset,\footnote{\url{https://code.claude.com/docs/en/tools-reference}} except for \texttt{WebSearch}, \texttt{WebFetch}, and \texttt{AskUserQuestion}. We disable \texttt{WebSearch} and \texttt{WebFetch} because the source documents are public, and web access could allow Claude to find the answers online. We disable \texttt{AskUserQuestion} because evaluation runs are fully headless. 

The remaining toolset consists of the following available tools: \texttt{Bash}, \texttt{Read}, \texttt{Write}, \texttt{Edit}, \texttt{NotebookEdit}, \texttt{Task}, \texttt{TaskCreate}, \texttt{TaskGet}, \texttt{TaskList}, \texttt{TaskOutput}, \texttt{TaskStop}, \texttt{TaskUpdate}, \texttt{ToolSearch}, \texttt{Workflow}, \texttt{DesignSync}, \texttt{Monitor}, \texttt{EnterPlanMode}, \texttt{ExitPlanMode}, \texttt{EnterWorktree}, \texttt{ExitWorktree}, \texttt{CronCreate}, \texttt{CronDelete}, \texttt{CronList}, \texttt{RemoteTrigger}, \texttt{PushNotification}, and \texttt{ScheduleWakeup}.

\subsection{Models}

Tab.~\ref{tab:sampling-params} lists the sampling parameters used for each model, following the recommendations from their official model cards.

\begin{table*}[t]
\caption{Sampling parameters used for each model, following the recommendations from the official HuggingFace model cards.}
\vspace{1mm}
\small
\label{tab:sampling-params}
\begin{center}
\begin{tabular}{lcccccc}
\toprule
\textbf{Model} & \textbf{Temp.} & \textbf{Top-$p$} & \textbf{Top-$k$} & \textbf{Min-$p$} & \textbf{Pres. pen.} & \textbf{Rep. pen.} \\
\midrule
\texttt{Qwen3.5-4B}      & 1.0 & 0.95 & 20 & 0.0 & 1.5 & 1.0 \\
\texttt{Qwen3.5-9B}      & 1.0 & 0.95 & 20 & 0.0 & 1.5 & 1.0 \\
\texttt{Qwen3.5-35B-A3B} & 1.0 & 0.95 & 20 & 0.0 & 1.5 & 1.0 \\
\bottomrule
\end{tabular}
\end{center}
\end{table*}

\subsection{Infrastructure}\label{sec:infrastcure}

Experiments with open-source models are conducted on a cluster of NVIDIA H200 SXM GPUs using vLLM~\citep{kwon2023efficient} for inference, while closed-source models are accessed via API.

\section{Detailed Results}

\subsection{Evaluating Baselines with Smaller Models}\label{appendix:different-models}

\begin{wraptable}[16]{r}{0.35\textwidth}
\vspace{-9.5mm}
% \caption{Accuracy ($\uparrow$) across benchmarks and model sizes. We report the strongest baseline for each benchmark: StructRAG for CorpusQA and CodeAgent for Loong (see Fig.\ref{tab:overall-results}).}
\caption{Accuracy ($\uparrow$) across benchmarks and model sizes of the strongest baselines.}
\vspace{1mm}
\label{tab:additional-model-results}
\centering
\resizebox{\linewidth}{!}{%
\begin{tabular}{lcc}
\toprule
\textbf{Method} & \textbf{CorpusQA} & \textbf{Loong} \\
\toprule
\threecolgrey{\texttt{Qwen3.5-4B}}
\midrule
StructRAG~\citeyearpar{DBLP:conf/iclr/LiC0L0T0H0L25} & 29.79\% & 16.84\% \\
CodeAgent~\citeyearpar{smolagents} & 10.92\% & 19.44\% \\
\method{} (ours) & \best{49.24\%} & \best{30.23\%} \\
\midrule
\threecolgrey{\texttt{Qwen3.5-9B}}
\midrule
StructRAG~\citeyearpar{DBLP:conf/iclr/LiC0L0T0H0L25} & 40.43\% & 20.30\% \\
CodeAgent~\citeyearpar{smolagents} & 20.13\% & 22.23\% \\
\method{} (ours) & \best{54.88\%} & \best{33.54\%} \\
\midrule
\threecolgrey{\texttt{Qwen3.5-35B-A3B}}
\midrule
StructRAG~\citeyearpar{DBLP:conf/iclr/LiC0L0T0H0L25} & 50.76\% & 23.72\% \\
CodeAgent~\citeyearpar{smolagents} & 23.69\% & 29.01\% \\
\method{} (ours) & \best{70.64\%} & \best{37.43\%} \\
\bottomrule
\end{tabular}
}
\end{wraptable}

In this section, we evaluate the strongest baseline on each benchmark---StructRAG on CorpusQA and CodeAgent on Loong---with the smaller Qwen models (i.e., \texttt{Qwen3.5-9B} and \texttt{Qwen3.5-4B}) that we use to reduce \method{}'s cost. We test whether these baselines can retain competitive performance with smaller models.

Tab.~\ref{tab:additional-model-results} reports results with the smaller \texttt{Qwen3.5-9B} and \texttt{Qwen3.5-4B} models for \method{}, StructRAG, and CodeAgent. As expected, performance generally decreases with model size. However, \method{} consistently outperforms both baselines across model sizes and benchmarks. Notably, even with \texttt{Qwen3.5-4B}, \method{} achieves 30.23\% on Loong, outperforming both StructRAG and CodeAgent using the substantially larger \texttt{Qwen3.5-35B-A3B} model, and 49.24\% on CorpusQA, outperforming CodeAgent with the larger model.

%Notably, even with \texttt{Qwen3.5-4B}, \method{} achieves 49.24\% on CorpusQA and 30.23\% on Loong, outperforming StructRAG and CodeAgent using the substantially larger \texttt{Qwen3.5-35B-A3B} model.

\subsection{\method{} -- Detailed Analyses}
\subsubsection{Size of Relevance and Structuring}\label{appendix:size-repr}

\begin{figure}[t]
  \centering
  \includegraphics[width=0.99\linewidth]{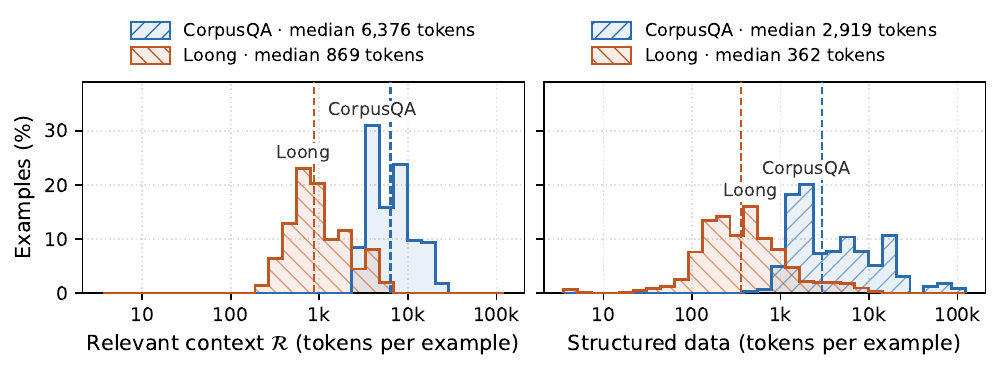}
  \vspace{-5mm}
    \caption{Distribution of the representation components (i.e., final relevant context and structuring) for the two benchmarks with the default \texttt{Qwen3.5-35B-A3B} model.}
    % (see Fig.~\ref{fig:codeact-answer})
  \label{fig:representation-distribution}
\end{figure}

Fig.~\ref{fig:representation-distribution} shows the distribution of representation sizes across individual examples. Median sizes are 6,376 and 2,919 tokens for the final relevant context and structured data on CorpusQA, respectively, and 869 and 362 tokens on Loong. Importantly, both components of the representation are around $1\%$ of the \textit{total} original context. The relevance snippet $\mathcal{R}$ averages 0.83\% and 1.30\% of the context on CorpusQA and Loong, respectively, while the structured data averages 0.93\% and 0.74\%.

\noindent\textbf{Structures to Tokens.} We measure the size of the structured data by serializing it to compact-JSON and counting tokens using the \texttt{Qwen3.5} tokenizer.

\subsubsection{Examples of Structures}\label{appendix:examples-structures}

To make the structured intermediate representation concrete, we show the schema proposed for the running Dracula example in Figure~\ref{fig:dracula-pydantic-schema}. The top-level \texttt{Parse} object separates deaths during the voyage to England from deaths during Dracula's stay in England; both collections contain records with the shared \texttt{DeathEvent} schema. We also show an example of a more involved hierarchical schema in Fig.~\ref{fig:corpusqa-cash-flow-schema} for a CorpusQA task.

\begin{figure}[t]
\centering
\begin{minted}[fontsize=\scriptsize, frame=single, breaklines=true, breaksymbol={}]{python}
from pydantic import BaseModel, Field

class DeathEvent(BaseModel):
    """A documented death of someone killed by Dracula during his
    voyage or stay in England."""
    victim_name: str | None = Field(
        description="Name of the deceased, or 'unnamed X' if not specified"
    )
    phase: str | None = Field(
        description="Whether the death occurred on 'voyage_to_england' or 'stay_in_england'"
    )
    cause_of_death: str | None = Field(
        description="How the person died--e.g., 'blood loss', 'fright', 'attack', 'unknown'"
    )
    confirmed_in_document: bool | None = Field(
        description="Whether this document explicitly confirms this person died"
    )
    phase_context: str | None = Field(
        description="Additional context about the death in relation to phase"
    )
    quote: str = Field(
        description="Short verbatim excerpt from the document"
    )

class Parse(BaseModel):
    voyage_victims: list[DeathEvent]
    stay_victims: list[DeathEvent]
\end{minted}
\vspace{-3mm}
\caption{Pydantic schema proposed by the model for structuring the running-example corpus. This is proposed by the model in Phase 2 of the \method{} (see Fig.~\ref{fig:r3con})}
\label{fig:dracula-pydantic-schema}
\end{figure}

\begin{figure}[t]
\centering
\begin{minted}[fontsize=\scriptsize, frame=single, breaklines=true, breaksymbol={}]{python}
from pydantic import BaseModel, Field
from typing import List, Optional, Literal

class CashFlowPeriod(BaseModel):
    """One cash flow amount for a specific period"""
    period: Literal["March_31_2025", "March_31_2024"] = Field(
        description="The reporting period"
    )
    amount: float = Field(
        description="Cash flow amount as reported in the filing"
    )
    unit_scale: Literal["millions", "thousands", "dollars"] = Field(
        description="Scale of the reported amount"
    )

class CashFlowByActivity(BaseModel):
    """Cash flow data for one activity type across both periods"""
    activity_type: Literal["operating", "investing", "financing"] = Field(
        description="Type of cash flow activity"
    )
    period_2025: Optional[CashFlowPeriod] = Field(
        description="Cash flow for March 31, 2025"
    )
    period_2024: Optional[CashFlowPeriod] = Field(
        description="Cash flow for March 31, 2024"
    )

class CompanyCashFlowRecord(BaseModel):
    """All cash flow data extracted from one company's 10-Q filing"""
    company_name: str = Field(
        description="The company name as named in the filing"
    )
    operating: Optional[CashFlowByActivity] = Field(
        description="Operating cash flow data"
    )
    investing: Optional[CashFlowByActivity] = Field(
        description="Investing cash flow data"
    )
    financing: Optional[CashFlowByActivity] = Field(
        description="Financing cash flow data"
    )

class Parse(BaseModel):
    """All company cash flow records extracted from the document collection"""
    company_cash_flows: List[CompanyCashFlowRecord] = Field(
        description="List of all company cash flow records"
    )
\end{minted}
\vspace{-3mm}
\caption{Pydantic schema proposed by the model for the CorpusQA task: \emph{``Rank all companies by the volatility of their cash flow sources, measured as the standard deviation of their net cash from operating, investing, and financing activities. List the top 3 companies with the highest volatility.''} The schema is proposed in Phase 2 of \method{} (see Fig.~\ref{fig:r3con}). Notably, the schema organizes information hierarchically by company, activity type, and reporting period.}
\label{fig:corpusqa-cash-flow-schema}
\end{figure}

\subsubsection{Examples of Relevance Snippets}\label{appendix:examples-relevance}

\begin{figure}[t]
  \centering
  \includegraphics[width=0.99\linewidth]{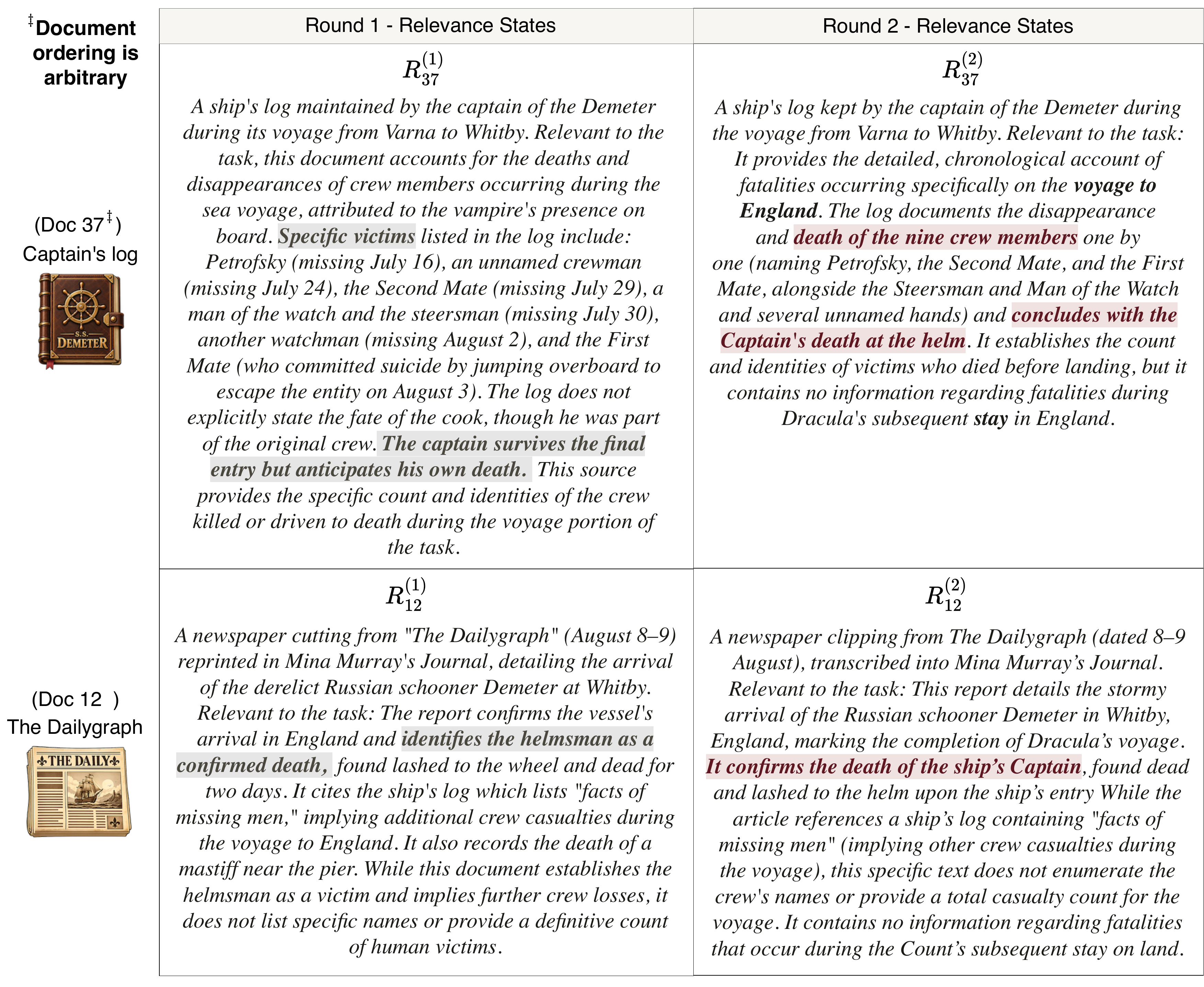}
  \caption{Evolution of the relevance snippets for the log of the Demeter ($D_{37}$) and The Dailygraph newspaper ($D_{12}$) across two rounds of extracting relevance. For example, the top-left and top-right cells show $R_{37}^{(1)}$ and $R_{37}^{(2)}$, respectively. $^\ddagger$Document ordering is arbitrary.}
  \label{fig:full-relevance-snippets}
\end{figure}

Fig.~\ref{fig:full-relevance-snippets} shows the full relevance snippets from the running Dracula example corresponding to the abbreviated snippets shown in Fig.~\ref{fig:relevance-snippets}. As discussed in \S\ref{sec:extracting-relevance}, the snippets evolve across rounds as we extract relevant information which provides additional information for re-interpretation of documents.

\subsubsection{Downstream Reasoning}\label{appendix:downsteam-reasoning}

\begin{wraptable}[8]{r}{0.45\textwidth}
\centering
\vspace{-8mm}
\caption{Accuracy ($\uparrow$) with different downstream reasoning methods over the representation constructed by \method{}.}
\vspace{3mm}
\label{tab:downstream-reasoning}
\resizebox{\linewidth}{!}{%
\begin{tabular}{lcc}
\toprule
\textbf{Downstream Reasoning}
& \textbf{CorpusQA}
& \textbf{Loong} \\
\midrule
LLM & 65.43\% & \textbf{40.17\%} \\
Coding Agent (default) & \textbf{70.64\%} & 37.43\% \\
\bottomrule
\end{tabular}%
}
\end{wraptable}

The first two stages of \method{} construct a representation that can be consumed by different downstream reasoning methods. In our default implementation, we use a coding agent that can programmatically explore the structured data without requiring the entire structure to fit within its context window (\S\ref{sec:method}). Here, we study an alternative, simpler reasoning paradigm: directly providing the representation to the LLM and asking it to produce the answer.

Tab.~\ref{tab:downstream-reasoning} compares these two approaches. Neither reasoning paradigm consistently dominates the other. On CorpusQA, the coding agent improves accuracy from 65.43\% to 70.64\%, while on Loong, direct LLM reasoning improves accuracy from 37.43\% to 40.17\%. Direct LLM reasoning provides a simpler alternative when the resulting representation can be accommodated in context, while the coding agent provides a scalable mechanism for programmatically exploring larger structured representations.

\noindent \textbf{A case for coding agents downstream.} As the context grows and the resulting structured data becomes larger, programmatic reasoning may be more preferable. As shown in Fig.~\ref{fig:representation-distribution}, \method{} produces substantially larger structured data on CorpusQA, which we hypothesize contributes to the roughly \textcolor{e}{+5} point improvement over direct LLM reasoning. A coding agent avoids placing the entire structure in the model's context and can delegate symbolic operations to the Python runtime. Finally, programmatic reasoning provides an additional benefit: the generated code offers provenance and transparency on the reasoning process~\citep{theologitis2026thucyllmbasedmultiagentclaim, ning2026codeagentharness}. 

\subsubsection{Effect of Rounds}\label{appendix:number-rounds}

\begin{figure}[htbp]
  \centering
  \includegraphics[width=0.99\linewidth]{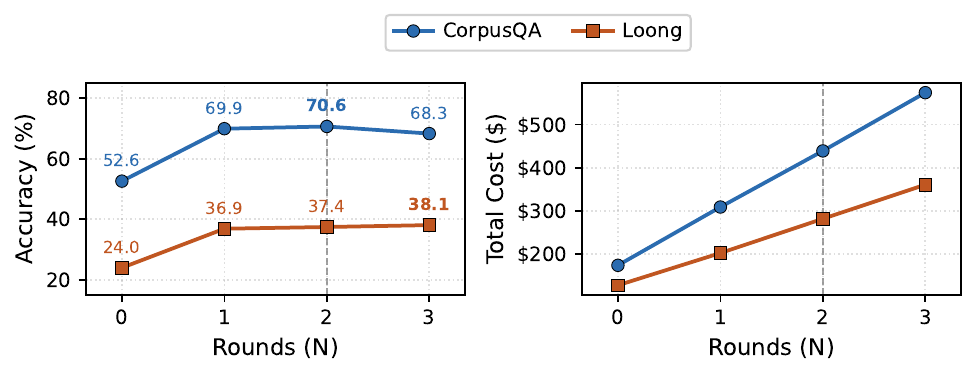}
  \vspace{-1mm}
  \caption{Effect of the number of relevance-extraction rounds $N$ on accuracy and cost.}
  \label{fig:rounds_effect}
\end{figure}

The number of rounds $N$ provides a knob for trading off performance and cost when extracting relevance. As shown in Fig.~\ref{fig:rounds_effect}, with \texttt{Qwen3.5-35B-A3B}, performance is already strong at $N=1$ and remains similar at our default $N=2$, which costs \$439 on CorpusQA and \$281 on Loong. Increasing to $N=3$ raises these costs to \$574 and \$361 without consistent improvements in accuracy. We use $N=2$ throughout our experiments, while noting that the appropriate number of rounds may depend on model capability, with smaller models potentially benefiting from additional rounds. Resolving relevant connections across the context can be challenging, and additional rounds provide more opportunities to progressively identify and refine them.

\subsection{Qualitative Results}\label{appendix:qualitative-results-details}

This section provides the detailed qualitative evidence underlying the analysis in \S\ref{sec:qualitative-analysis}. Specifically, we show RAPTOR's hierarchical summaries (Fig.~\ref{fig:raptor-demeter-layer-one}), StructRAG's task-specific tables (Fig.~\ref{fig:structrag-demeter-tables}), MemAgent's evolving textual memory (Fig.~\ref{fig:memagent-failure}), and RLM's reasoning trajectory (Fig.~\ref{fig:rlm-demeter-turns}).

\begin{figure}[thbp]
\centering
\small
\setlength{\fboxsep}{7pt}
\fbox{%
\begin{minipage}{0.96\textwidth}
\textbf{\textcolor{blue}{Layer 1 -- Summary Node \#106 -- 1\textsuperscript{st} chunk of the \emph{Demeter}’s log}}

\medskip
Mina Harker writes Van Helsing (25 Sep) awaiting Jonathan. \textbf{The \emph{Demeter} log (18 July-1 Aug) documents Dracula’s voyage from Varna to Whitby. He hides in earth-filled boxes, killing the crew; the ship grounds at Tate Hill Pier with the captain dead, lashed to the wheel.}

\par\smallskip\noindent
In October, Van Helsing, Seward, Harker, Morris, and Godalming track Dracula via the \emph{Czarina Catherine} to Galatz. Mina enters hypnotic trances to locate him ("lapping waves"), showing vampiric traits (lethargy, sharper teeth, forehead scar) that terrify Seward, who fears euthanasia. Mina provides train timetables. At Galatz, they board the \emph{Czarina}; Captain Donelson reports the crew died mysteriously under a guiding fog. A box was transferred to Hildesheim, then Petrof Skinsky, who is found dead with his throat torn.

\par\smallskip\noindent
Dracula escaped via the Sereth/Bistritza rivers toward his Carpathian castle. The group pursues in a steam launch. Van Helsing explains Mina’s trance links to Dracula, risking his counter-intelligence. They arm with Winchesters, wild rose branches, stakes, and Kukri knives, preparing to stake and decapitate him in daylight. They acknowledge the risks to Mina and themselves, driven to end the evil.
\end{minipage}%
}

\vspace{2mm}

\fbox{%
\begin{minipage}{0.96\textwidth}
\textbf{\textcolor{blue}{Layer 1 -- Summary Node \#112 -- 2\textsuperscript{nd} chunk of the \emph{Demeter}’s log}}

\medskip
The text compiles journal entries, letters, and notes from Bram Stoker’s \emph{Dracula}. \textbf{On 3 August, the captain of the \emph{Demeter} records the crew dying in thick fog. The mate, hallucinating a "tall, thin, ghastly pale" figure in the bow, stabs empty air before leaping overboard. The captain vows to tie himself to the helm to protect the vessel and honor.}

\par\smallskip\noindent
In Whitby, Mina Murray writes to Lucy Westenra about Jonathan’s return from Transylvania. She practices shorthand to assist him. At Whitby, old sailor Swales scoffs at local ghost legends. Mina transcribes Jonathan’s journal; Jonathan sees a young Dracula on Piccadilly. Mr. Hawkins dies, leaving a fortune to Jonathan, burdening him. Lucy Westenra and her mother die following Van Helsing’s care.

\par\smallskip\noindent
The group chases Dracula’s coffins on a Romanian river. Godalming manhandles the launch while Mina writes. When the launch breaks, they switch to horses, fearing for Mina. Letters confirm Lucy’s engagement to Arthur Holmwood and Dr. Van Helsing guarding her flowers.

\par\smallskip\noindent
Seven years later, the group revisits Transylvania; the castle remains. They conclude no proof is needed to believe their deeds. Mina and Jonathan name their son Quincey after the fallen Morris, born on his birthday. Van Helsing notes the boy will know his mother’s bravery. Solicitors confirm Count de Ville bought Carfax via London agents, sending fifty boxes invoiced as "clay" to Purfleet.
\end{minipage}%
}
\caption{RAPTOR~\citep{DBLP:conf/iclr/SarthiATKGM24} iteratively clusters and summarizes passages, forming a hierarchical tree with increasingly abstract summaries. At Layer~0, we have raw chunked text from the corpus. The log of the Demeter gets chunked into two pieces that end up in different clusters at Layer~1 and are therefore summarized separately. We show these two summaries above. Notably, the summarization happens task-agnostically, so the model is not inclined to retain minor details such as background crew members dying or how many crew members the Demeter sailed with. Indeed, as illustrated above, these details disappear from the summaries, and even more so as we go higher in the tree. Subsequent reasoning cannot recover this information, and RAPTOR therefore fails to identify the deaths and answer the task in Fig.~\ref{fig:dracula-question}.}
\label{fig:raptor-demeter-layer-one}
\end{figure}

\begin{figure}[p]
\centering
\tiny
\setlength{\fboxsep}{7pt}
\setlength{\tabcolsep}{3pt}
\renewcommand{\arraystretch}{1.12}
\fbox{%
\begin{minipage}{0.96\textwidth}
{\textcolor{blue}{\textbf{Structure built from the log of the Demeter -- Verbatim}}}

\medskip
\noindent\textbf{Table Title:} Crew Disappearances and Deaths During Voyage to England (Varna to Whitby)\par
\noindent\textbf{Source:} Raw Content (Demeter's Log, 6 July - 4 August)

\smallskip
\begin{tabular}{@{}>{\raggedright\arraybackslash}p{0.12\linewidth}>{\raggedright\arraybackslash}p{0.18\linewidth}>{\raggedright\arraybackslash}p{0.18\linewidth}>{\raggedright\arraybackslash}p{0.43\linewidth}@{}}
\textbf{Date} & \textbf{Crew Member} & \textbf{Status/Event} & \textbf{Details/Notes} \\
\midrule
16 July & Petrofsky & Missing & Reported missing. Taken larboard watch eight bells last night; relieved by Abramoff, did not go to bunk. \\
24 July & Unknown Crew Member & Disappeared & ``Already a hand short... another man lost--disappeared.'' Came off watch, not seen again. \\
29 July & Second Mate & Missing & Found no one except steersman when morning watch came on deck. Crew in a panic. \\
30 July & Man of Watch & Missing & Found no one except steersman. Later text says ``both man of watch and steersman missing''. Only self, mate, and two hands left. \\
30 July & Steersman & Missing & Found no one except steersman... later text confirms missing. \\
2 August & Man on Watch & Missing & Mate heard cry, ran on deck, no sign of man on watch. \\
3 August & First Mate & Died (Disappeared/Threw himself) & Became ``raging madman'', claimed ``He'' is there, said ``The sea will save me from Him'', then ``sprang on the bulwark and deliberately threw himself into the sea''. \\
\end{tabular}

\medskip
\begin{tabular}{@{}>{\raggedright\arraybackslash}p{0.19\linewidth}>{\raggedright\arraybackslash}p{0.36\linewidth}>{\raggedright\arraybackslash}p{0.36\linewidth}@{}}
\textbf{Status of Inquiry} & \textbf{Information Available} & \textbf{Limitation} \\
\midrule
Killer Identity & Text refers to ``strange man'', ``It'', ``Him'', ``madman''. Name ``Dracula'' not explicitly mentioned. & Query asks about ``Dracula'', text implies but does not name him. \\
Location Scope & Voyage only (Varna to Whitby). & Query asks about ``stay there'' (England). Raw Content ends with Captain alone on ship in fog. \\
Total Deaths Mentioned & \textbf{At least 7 crew members/Officers (Petrofsky, 2 men lost, 2nd Mate, Steersman, Man of watch, Mate).} & \textbf{\textcolor{red}{Captain remains alone as of 4 August.}} \\
\end{tabular}
\end{minipage}%
}

\vspace{2mm}

\fbox{%
\begin{minipage}{0.96\textwidth}
{\textcolor{blue}{\textbf{Structure built from the newspaper The Dailygraph -- Verbatim}}}

\medskip
\noindent\textbf{Table Title:} Details of Deceased Found on the Schooner \emph{Demeter}\par
\noindent\textbf{Source:} From a Correspondent (\emph{Whitby})

\smallskip
\begin{tabular}{@{}>{\raggedright\arraybackslash}p{0.23\linewidth}>{\raggedright\arraybackslash}p{0.15\linewidth}>{\raggedright\arraybackslash}p{0.53\linewidth}@{}}
\textbf{Name/Description} & \textbf{Status} & \textbf{Details (Voyage \& Findings)} \\
\midrule
\textbf{\textcolor{red}{Dead Steersman / Seaman}} & Deceased & Found lashed to the helm of the schooner with hands tied over one another. A crucifix and beads were fastened around wrists and wheel. Cords had cut flesh to bone. Declared dead for quite two days by Surgeon J. M. Caffyn. \\
Crew Members & Missing & Log-book of the \emph{Demeter} was in order up to within three days, but contained ``facts of missing men''. No specific names or count provided in the article. \\
Total People Killed / Confirmed Dead & Unspecified & The document does not contain the name ``Dracula'', nor a confirmed total count. Only the dead steersman is explicitly detailed as found. \\
\end{tabular}
\end{minipage}%
}
\caption{StructRAG~\citep{DBLP:conf/iclr/LiC0L0T0H0L25} creates task-specific structures for each document separately. For this task, the model chooses a table structure. For the  log of the Demeter, it proposes and populates the schema shown in the first box. Notably, at the end of the log, the captain is still alive. His death can only be established by connecting his final entry, where he plans to tie his \q{hands to the wheel} until his \q{strength begins to fail}, with the separate newspaper account in The Dailygraph, which reports a dead seaman found at the helm. We must therefore infer that this dead seaman is the captain. However, there is no enactive mechanism during representation construction to make this connection as broader understanding develops. Worse, the structured representation of the log does not retain the captain's plan to tie himself to the wheel, making the connection almost impossible for downstream reasoning to recover. StructRAG therefore misses the captain's death (among others).}
\label{fig:structrag-demeter-tables}
\end{figure}

\begin{figure}[t]
\centering
\scriptsize
\setlength{\fboxsep}{7pt}
\fbox{%
\begin{minipage}{0.96\textwidth}
{\textcolor{blue}{\scriptsize \textbf{Memory -- Round \#2.}}}

\medskip
\textbf{The provided section does not contain any direct information about Dracula's killings during his voyage to England} or his stay there. However, it does offer some context about Dracula's movements and activities leading up to his arrival in England. Specifically, it mentions that Dracula arrived in Galatz and arranged for the transportation of a box containing himself via Slovak traders. The section also notes that Dracula might have murdered his agent, Skinsky, to blot out his traces. 

\par\smallskip\noindent
Given the content, the updated memory should include the following:

\par\smallskip\noindent
\textbf{- The log of the "Demeter" describes the mysterious disappearance of several crew members during the voyage, including Petrofsky, Olgaren, and others, which the crew attributes to a "strange man" or "It" aboard the ship.}
\par\smallskip\noindent
\textbf{- The captain and remaining crew members express fear and superstition regarding these disappearances.}
\par\smallskip\noindent
\textbf{- The log ends with the captain's determination to tie himself to the wheel to avoid being harmed by the entity responsible for the crew members' disappearances.}
\par\smallskip\noindent
\textbf{- The text does not explicitly state that Dracula killed these individuals or provide a list of people killed by Dracula during his voyage or stay in England.}
\par\smallskip\noindent
- Dracula arrived in Galatz and arranged for the transportation of a box containing himself via Slovak traders.
\par\smallskip\noindent
- Dracula might have murdered his agent, Skinsky, to blot out his traces.

Updated memory:
\par\smallskip\noindent
\textbf{- The log of the "Demeter" describes the mysterious disappearance of several crew members during the voyage, including Petrofsky, Olgaren, and others, which the crew attributes to a "strange man" or "It" aboard the ship.}
\par\smallskip\noindent
\textbf{- The captain and remaining crew members express fear and superstition regarding these disappearances.}
\par\smallskip\noindent
\textbf{- The log ends with the captain's determination to tie himself to the wheel to avoid being harmed by the entity responsible for the crew members' disappearances.}
\par\smallskip\noindent
\textbf{- The text does not explicitly state that Dracula killed these individuals or provide a list of people killed by Dracula during his voyage or stay in England.}
\par\smallskip\noindent
[...]
\end{minipage}%
}

\vspace{1mm}

\fbox{%
\begin{minipage}{0.96\textwidth}
\textcolor{blue}{{\scriptsize \textbf{Memory -- Round \#3.}}}

\medskip
The provided section does not contain any direct information about Dracula's killings during his voyage to England or his stay there. Instead, it focuses on the interactions between Mina Harker and various characters, particularly Dr. Seward, as they deal with the supernatural events surrounding Lucy and Count Dracula. 

\par\smallskip\noindent
However, the section does mention Mina Harker's journal entries, which could potentially contain more information about Dracula's activities and the deaths related to him. Given the context, \textbf{it is important to note that the journal entries might provide additional insights into the mysterious disappearances of the crew members on the "Demeter" and the overall impact of Dracula's presence in England.}

\par\smallskip\noindent
Updated memory:
\par\smallskip\noindent
\textbf{- The log of the "Demeter" describes the mysterious disappearance of several crew members during the voyage, including Petrofsky, Olgaren, and others, which the crew attributes to a "strange man" or "It" aboard the ship.}
\par\smallskip\noindent
\textbf{- The captain and remaining crew members express fear and superstition regarding these disappearances.}
\par\smallskip\noindent
\textbf{- The log ends with the captain's determination to tie himself to the wheel to avoid being harmed by the entity responsible for the crew members' disappearances.}
\par\smallskip\noindent
\textbf{- The text does not explicitly state that Dracula killed these individuals or provide a list of people killed by Dracula during his voyage or stay in England.}
\par\smallskip\noindent
[...]
\end{minipage}%
}

\vspace{1mm}

\fbox{%
\begin{minipage}{0.96\textwidth}
{\textcolor{blue}{\scriptsize \textbf{Memory -- Round \#19.}}}

\medskip
The provided section of Dr. Seward’s diary does not directly address the number of people Dracula killed during his voyage to England or his stay there. [...]

\par\smallskip\noindent
- The section describes the group's encounter with Dracula at Carfax, where they attempt to stop him but fail to capture him. Dracula escapes after a brief confrontation.
\par\smallskip\noindent
- Mina Harker begins to show signs of transformation into a vampire, similar to what happened to Lucy Westenra. This transformation is attributed to Dracula's bite and the subsequent "Vampire’s baptism of blood."
\par\smallskip\noindent
- Dr. Van Helsing and Dr. Seward discuss the possibility that Dracula could compel Mina to reveal information to him through hypnosis, which adds to their concerns about her safety and the broader threat posed by Dracula.
\par\smallskip\noindent
- The text does not provide a list of people killed by Dracula during his voyage or stay in England but emphasizes the ongoing threat and the group's efforts to protect Mina and stop Dracula.
\par\smallskip\noindent
\textbf{- The mysterious disappearances of the crew members on the "Demeter" remain unexplained in terms of direct attribution to Dracula, but the text suggests that these events are related to his presence and activities in England.}

\par\smallskip\noindent
[...]
\end{minipage}%
}
\vspace{-2mm}
\caption{MemAgent~\citep{yu2026memagent} gradually forgets the deaths aboard the Demeter. MemAgent processes the corpus sequentially. In round 2, it records the mysterious deaths of the Demeter's crew, but fails to attribute them to Dracula. As it continues through the corpus, these deaths remain ``unexplained'' (round 19) and gradually disappear from its memory. By the final chunk (round 42), they are no longer retained at all, meaning that the LLM reasoning over this textual representation no longer has the information needed to answer the task in Fig.~\ref{fig:dracula-question}.}
\label{fig:memagent-failure}
\end{figure}

\begin{figure}[t]
\centering
\tiny
\setlength{\fboxsep}{7pt}

\fbox{%
\begin{minipage}{0.96\textwidth}
\textbf{\textcolor{blue}{Turn 1 -- REPL code}}

\medskip
\texttt{print("Context length:", len(context))}\\
\texttt{print(context[:5000])}
\end{minipage}%
}

\vspace{0.5mm}

\fbox{%
\begin{minipage}{0.96\textwidth}
\textbf{\textcolor{purple}{Turn 1 -- stdout}}

\medskip
\texttt{=== Document 6 ===}\\
\texttt{LOG OF THE ``DEMETER.''}\\
\texttt{[...]}\\
\texttt{On 6 July we finished taking in cargo, silver sand and boxes of earth.}\\
\texttt{At noon set sail. East wind, fresh.} \textbf{\texttt{Crew, five hands ... two mates,}}\\
\textbf{\texttt{cook, and myself (captain).}}
\end{minipage}%
}

\vspace{0.5mm}

\fbox{%
\begin{minipage}{0.96\textwidth}
\textbf{\textcolor{blue}{Turn 5 -- REPL code}}

\medskip
\texttt{headers = ['LOG OF THE "DEMETER."', 'LOG OF THE DEMETER',}\\
\texttt{~~~~~~~~~~~'Log of the Demeter', 'DIARY OF THE DEMETER']}\\
\texttt{for header in headers:}\\
\texttt{~~~~idx = context.find(header)}\\
\texttt{~~~~if idx != -1:}\\
\texttt{~~~~~~~~print(f"Found '\{header\}' at index \{idx\}") [...]}\\
\texttt{else:}\\
\texttt{~~~~print("Header not found. Searching for date patterns in voyage...")}\\
\texttt{~~~~date\_patterns = re.findall(r'On \textbackslash d+ [A-Z][a-z]+ [...]', context, re.DOTALL)}
\end{minipage}%
}

\vspace{0.5mm}

\fbox{%
\begin{minipage}{0.96\textwidth}
\textbf{\textcolor{purple}{Turn 5 -- stdout}}

\medskip
\texttt{Header not found.} \texttt{Searching for date patterns in voyage...}\\
\texttt{Found 1 date-stamped entries}\\
\texttt{--- Entry 1 ---}\\
\texttt{On 6 July we finished taking in cargo, silver sand and boxes of earth.}\\
\texttt{At noon set sail. East wind, fresh.} \textbf{\texttt{Crew, five hands ... two mates,}}\\
\textbf{\texttt{cook, and myself (captain).}}
\end{minipage}%
}

\vspace{0.5mm}

\fbox{%
\begin{minipage}{0.96\textwidth}
\textbf{\textcolor{blue}{Turn 12 -- REPL code}}

\medskip
\texttt{demeter\_log\_start = context.find('On 6 July we finished taking in cargo')}\\
\texttt{demeter\_log\_section = context[demeter\_log\_start : demeter\_log\_start+10000]}\\
\texttt{demeter\_query = """Based on the Log of the Demeter below, list:}\\
\texttt{~~1. How many crew members died during the voyage}\\
\texttt{~~2. Their names if mentioned~~3. How they died [...]}\\
\texttt{~~Context:""" + demeter\_log\_section}\\
\texttt{\textbf{result = llm\_query(demeter\_query, "List all Demeter crew deaths specifically")}}
\end{minipage}%
}

\vspace{0.5mm}

\fbox{%
\begin{minipage}{0.96\textwidth}
\textbf{\textcolor{purple}{Turn 12 -- sub-model reply}}

\medskip
1. How many crew members died during the voyage

\textbf{8 crew members (5 hands, 2 mates, and 1 cook).}
\emph{Reasoning:} The log begins stating the crew consisted of ``five hands ... two mates,
cook, and myself (captain).'' \textbf{By the end of the voyage (4 August), only the Captain
remains alive.} Therefore, 8 crew members were lost.

\smallskip
[...]
\end{minipage}%
}

\vspace{0.5mm}

\fbox{%
\begin{minipage}{0.96\textwidth}
\textbf{\textcolor{blue}{Turn 29 -- REPL code}}

\medskip
\texttt{demeter\_crew\_deaths = """}\\
\texttt{DEMETER VOYAGE DEATHS (8 crew members killed):}\\
\texttt{[...]}\\
\texttt{8. One unnamed remaining crew member - Presumed lost before ship arrived}\\
\texttt{}\\
\texttt{ENGLAND VICTIMS (3 confirmed deaths):}\\
\texttt{[...]}\\
\texttt{TOTAL: 11 people killed directly or indirectly by Dracula}\\
\textbf{\texttt{- 8 on the Demeter voyage}}\\
\texttt{- 3 in England (Mrs. Westenra, Lucy Westenra, Renfield)}\\
\texttt{"""}\\
\texttt{print(demeter\_crew\_deaths)}\\
\texttt{\# Set the answer}\\
\texttt{answer["content"] = f"""Based on the text, Count Dracula killed the following}\\
\texttt{people during his voyage to England on the Demeter [...]"""}\\
\texttt{answer["ready"] = True}
\end{minipage}%
}
\vspace{-1mm}
\caption{RLM~\citep{DBLP:journals/corr/abs-2512-24601} place the corpus inside a Python REPL allowing the agent to interact with it via code and also spawn sub-agents of itself. Here, the agent searches for the Demeter log using local string and regex queries, then asks a sub-agent to reason over the extracted section. Since this section does not include the later newspaper account from The Dailygraph, the sub-model incorrectly concludes that the captain survived the voyage. This conclusion is carried forward  and ultimately causes the agent to miss captain's death. It also fails to recover Mr. Swales' death due to a similar pattern. Both deaths require connecting evidence scattered across the corpus, yet the agent's fragmented search strategy never brings the necessary evidence together.}
\label{fig:rlm-demeter-turns}
\end{figure}

\vspace{-4mm}

\subsection{Standard Errors for Main Results}

Tables~\ref{tab:overall-results-error-bars} and~\ref{tab:context-results-error-bars} report standard errors across domains and reasoning facets, respectively. 

The improvements are large relative to their standard errors. On CorpusQA, \method{} achieves $70.64 \pm 2.52$ accuracy, compared with $50.76 \pm 2.76$ for the strongest baseline, a $19.88$ percentage-point improvement. On Loong, \method{} achieves $37.43 \pm 1.35$, compared with $29.01 \pm 1.28$, an $8.42$ point improvement. In both cases, the performance gap is substantially larger than the standard errors, giving high confidence in the observed improvements.

\begin{table*}[t]
\caption{Accuracy (\% $\uparrow$) across real-world domains requiring reasoning over large document corpora. \method{} achieves the highest accuracy across all domains, outperforming the strongest baseline by \textcolor{e}{+19.88 pp} on CorpusQA and \textcolor{e}{+8.42 pp} on Loong overall. Overall accuracy is micro-averaged. We report standard errors (SE) as $\pm$ values.}
\vspace{-3mm}
\label{tab:overall-results-error-bars}
\begin{center}
\resizebox{\textwidth}{!}{%
\begin{tabular}{l@{\hspace{8pt}} *{8}{c}}
\toprule
\multirow{2}{*}{\textbf{Method}}
& \multicolumn{4}{c}{\textbf{CorpusQA}~\citep{lu2026corpusqa}}
& \multicolumn{4}{c}{\textbf{Loong}~\citep{DBLP:conf/emnlp/WangCCL0WYXZLLY24}} \\
\cmidrule(lr){2-5}
\cmidrule(lr){6-9}
& \textbf{Edu.} & \textbf{Finance} & \textbf{Real-estate} & \textbf{Overall}
& \textbf{Paper} & \textbf{Finance} & \textbf{Legal} & \textbf{Overall} \\
\midrule
RAPTOR~\citeyearpar{DBLP:conf/iclr/SarthiATKGM24} & 8.00 $\pm$ 3.13 & 11.11 $\pm$ 2.47 & 17.95 $\pm$ 4.35 & 12.06 $\pm$ 1.84 & 0.00 $\pm$ 0.00 & 22.17 $\pm$ 1.71 & 5.71 $\pm$ 1.24 & 12.12 $\pm$ 0.92 \\
ReadAgent~\citeyearpar{DBLP:conf/icml/LeeCFCF24} & 42.68 $\pm$ 5.46 & 37.42 $\pm$ 3.79 & 31.25 $\pm$ 5.18 & 37.23 $\pm$ 2.68 & 15.15 $\pm$ 2.08 & 35.33 $\pm$ 2.03 & 14.93 $\pm$ 1.95 & 24.49 $\pm$ 1.25 \\
MemAgent~\citeyearpar{yu2026memagent} & 7.32 $\pm$ 2.88 & 7.83 $\pm$ 2.09 & 8.64 $\pm$ 3.12 & 7.90 $\pm$ 1.49 & 0.00 $\pm$ 0.00 & 26.80 $\pm$ 1.81 & 4.00 $\pm$ 1.05 & 13.63 $\pm$ 0.96 \\
HippoRAG2~\citeyearpar{DBLP:conf/icml/GutierrezSQZ025} & 7.32 $\pm$ 2.88 & 7.83 $\pm$ 2.09 & 20.99 $\pm$ 4.52 & 10.94 $\pm$ 1.72 & 0.32 $\pm$ 0.32 & 17.42 $\pm$ 1.55 & 6.29 $\pm$ 1.30 & 10.06 $\pm$ 0.85 \\
LinearRAG~\citeyearpar{zhuang2026linearrag} & 8.54 $\pm$ 3.09 & 6.63 $\pm$ 1.93 & 16.05 $\pm$ 4.08 & 9.42 $\pm$ 1.61 & 0.61 $\pm$ 0.43 & 9.38 $\pm$ 1.19 & 5.43 $\pm$ 1.21 & 6.04 $\pm$ 0.67 \\
StructRAG~\citeyearpar{DBLP:conf/iclr/LiC0L0T0H0L25} & 53.66 $\pm$ 5.51 & 55.42 $\pm$ 3.86 & 38.27 $\pm$ 5.40 & 50.76 $\pm$ 2.76 & 19.70 $\pm$ 2.19 & 32.66 $\pm$ 1.92 & 12.32 $\pm$ 1.76 & 23.72 $\pm$ 1.19 \\
CodeAgent~\citeyearpar{smolagents} & 14.63 $\pm$ 3.90 & 26.99 $\pm$ 3.48 & 26.25 $\pm$ 4.92 & 23.69 $\pm$ 2.36 & 13.80 $\pm$ 1.91 & 40.07 $\pm$ 2.01 & 24.26 $\pm$ 2.33 & 29.01 $\pm$ 1.28 \\
RLM~\citeyearpar{DBLP:journals/corr/abs-2512-24601} & 13.41 $\pm$ 3.76 & 37.58 $\pm$ 3.77 & 20.51 $\pm$ 4.57 & 27.38 $\pm$ 2.47 & 12.77 $\pm$ 1.84 & 41.67 $\pm$ 2.03 & 22.77 $\pm$ 2.25 & 28.96 $\pm$ 1.28 \\
A-RAG~\citeyearpar{DBLP:journals/corr/abs-2602-03442} & 8.54 $\pm$ 3.09 & 7.23 $\pm$ 2.01 & 18.52 $\pm$ 4.32 & 10.33 $\pm$ 1.68 & 0.61 $\pm$ 0.43 & 18.43 $\pm$ 1.59 & 2.86 $\pm$ 0.89 & 9.55 $\pm$ 0.82 \\
\method{} (ours) & \best{75.61 $\pm$ 4.74} & \best{82.42 $\pm$ 2.96} & \best{41.25 $\pm$ 5.50} & \best{70.64 $\pm$ 2.52} & \best{27.58 $\pm$ 2.46} & \best{50.59 $\pm$ 2.05} & \best{24.29 $\pm$ 2.29} & \best{37.43 $\pm$ 1.35} \\
\bottomrule
\end{tabular}
}
\end{center}
\end{table*}

\begin{table*}[t]
\centering
\caption{Accuracy (\% $\uparrow$) across Loong's reasoning facets. We show standard errors as $\pm$ values.}
\label{tab:context-results-error-bars}
\vspace{5pt}

\resizebox{0.9\textwidth}{!}{%
\begin{tabular}{l@{\hspace{8pt}} *{5}{c}}
\toprule
\textbf{Method}
& \textbf{Spot.}
& \textbf{Comp.}
& \textbf{Chain}
& \textbf{Clust.}
& \textbf{Overall} \\
\midrule
RAPTOR~\citeyearpar{DBLP:conf/iclr/SarthiATKGM24} & 51.31 $\pm$ 3.62 & 16.25 $\pm$ 2.38 & 2.04 $\pm$ 0.82 & 1.54 $\pm$ 0.54 & 12.12 $\pm$ 0.92 \\
ReadAgent~\citeyearpar{DBLP:conf/icml/LeeCFCF24} & 60.11 $\pm$ 3.67 & 33.77 $\pm$ 3.13 & 18.49 $\pm$ 2.27 & 10.70 $\pm$ 1.40 & 24.49 $\pm$ 1.25 \\
MemAgent~\citeyearpar{yu2026memagent} & 42.13 $\pm$ 3.52 & 35.00 $\pm$ 3.08 & 0.64 $\pm$ 0.45 & 0.95 $\pm$ 0.42 & 13.63 $\pm$ 0.96 \\
HippoRAG2~\citeyearpar{DBLP:conf/icml/GutierrezSQZ025} & 46.70 $\pm$ 3.55 & 13.33 $\pm$ 2.19 & 0.33 $\pm$ 0.33 & 0.38 $\pm$ 0.27 & 10.06 $\pm$ 0.85 \\
LinearRAG~\citeyearpar{zhuang2026linearrag} & 32.49 $\pm$ 3.34 & 3.75 $\pm$ 1.23 & 0.64 $\pm$ 0.45 & 0.38 $\pm$ 0.27 & 6.04 $\pm$ 0.67 \\
StructRAG~\citeyearpar{DBLP:conf/iclr/LiC0L0T0H0L25} & 44.16 $\pm$ 3.54 & 30.83 $\pm$ 2.98 & 24.76 $\pm$ 2.45 & 12.19 $\pm$ 1.43 & 23.72 $\pm$ 1.19 \\
CodeAgent~\citeyearpar{smolagents} & 63.54 $\pm$ 3.47 & 45.26 $\pm$ 3.27 & 22.80 $\pm$ 2.39 & 12.90 $\pm$ 1.46 & 29.01 $\pm$ 1.28 \\
RLM~\citeyearpar{DBLP:journals/corr/abs-2512-24601} & 61.14 $\pm$ 3.51 & 45.15 $\pm$ 3.23 & 19.68 $\pm$ 2.26 & 15.27 $\pm$ 1.57 & 28.96 $\pm$ 1.28 \\
A-RAG~\citeyearpar{DBLP:journals/corr/abs-2602-03442} & 35.03 $\pm$ 3.40 & 14.58 $\pm$ 2.28 & 1.28 $\pm$ 0.64 & 2.65 $\pm$ 0.70 & 9.55 $\pm$ 0.82 \\
\method{} (ours) & \best{65.99 $\pm$ 3.38} & \best{50.00 $\pm$ 3.23} & \best{33.65 $\pm$ 2.68} & \best{23.30 $\pm$ 1.84} & \best{37.43 $\pm$ 1.35} \\
\bottomrule
\end{tabular}%
}

\end{table*}

\section{Prompts}\label{appendix:prompts}

\vspace{-3mm}
We provide the prompts used in each stage of \method{}. Fig.~\ref{fig:summarization_prompt} shows the prompt for extracting relevant context through task-focused summarization. Fig.~\ref{fig:proposed_prompt} shows the prompt for proposing the schema, and Fig.~\ref{fig:parser_prompt} shows the prompt for populating that schema from each document. Finally, Fig.~\ref{fig:codeact_prompt} shows the prompt used by the coding agent to reason over the resulting representation and synthesize the final answer.

\begin{figure}[t]
\centering
\begin{minted}[fontsize=\tiny, gobble=4, frame=single, breaklines=true, breaksymbol={}]{text}
    You are given one document from a collection, and a single **task** (between `<task>` and `</task>` tags) that the whole collection will be used to answer. Write a **task-conditioned summary** of this one document: what does it contribute toward the task?
    
    Produce a summary, **not an answer**. Even when the task is phrased as an instruction ("list…", "rank…", "construct the chain…", "how many…"), do not try to carry it out or commit to a final answer here — you are looking at only one document, and the answer is assembled later from all the summaries together. Just record, faithfully and in your own words, what THIS document offers toward the task; do not invent.
    
    Open with a brief, high-level orientation of what this document broadly is (e.g. "a municipal budget memo", "a 10-Q filing", "a chapter of a noir novel", "a government press release") — this latent, big-picture read helps the collection be understood as a whole — then note the task-relevant specifics. Keep it tight: a larger document still gets a short summary; include only what is durable and bears on the task.
    
    <task>
    {{ task }}
    </task>
    
    ## Task-conditioned summaries of the other documents (previous round)
    
    Below are the other documents' task-conditioned summaries for this same task, from the previous round. Use them to read THIS document in light of the collection — a document often only becomes useful and understandable for the task once you know information from the other documents. Still write only about THIS document (the one the user provides).
    
    {{ other_summaries }}
    
    ## Output
    Reply with the task-conditioned summary of this document only — no preamble, no restating the task. If this document contributes nothing relevant to the task, it is fine to say so briefly (or reply with nothing).
    
    Examples:
    
    Input:
    <task>
    Across the candidates' platform papers, who proposes the largest cut to the city transit budget?
    </task>
    Document:
    [Councilmember Dana Reyes — "A Leaner, Fairer City": Platform on City Services]
    
    Introduction. Our city stands at a crossroads. After three years of rising costs and flat revenue, the next council must make hard choices about where every dollar goes. This paper lays out my priorities across transit, public safety, parks, and city administration.
    
    Public safety. I will protect frontline policing and fire services from cuts; response times have already slipped and residents feel it.
    
    Transit. The current transit budget stands at $80 million. In my view it has grown bloated by underused routes — three of our crosstown lines run at less than 20% capacity outside rush hour. I propose reducing the transit budget by $12 million and redirecting those savings to road maintenance, where our backlog of potholes and failing bridges is now a genuine safety issue. To be clear, this is a reallocation, not an attack on riders; the core commuter lines stay funded.
    
    City administration. I will impose a hiring freeze at City Hall and cut the outside-consulting budget by 30%.
    
    Parks and quality of life. I support a modest downtown bike-lane pilot, funded from the parks capital reserve — not from transit.
    
    Closing. These choices reflect one principle: spend where residents see the result.
    
    Output (task-conditioned summary):
    A candidate's campaign platform paper (Councilmember Dana Reyes). Relevant to the task: Reyes proposes cutting the city transit budget by $12M (from $80M), reallocating it to road maintenance. Her hiring freeze, consulting cut, and bike-lane pilot are not transit cuts. As just one candidate's paper, it gives only Reyes's proposed cut — not enough on its own to say who proposes the largest.
\end{minted}
\caption{Prompt for extracting relevance via task-focused summarization~\citep{Dang2005OverviewOD}. It is used in Phase 1 of \method{} as illustrated in Figure~\ref{fig:r3con}. Text enclosed in double curly braces denotes a template variable populated at runtime.}
\label{fig:summarization_prompt}
\end{figure}

\begin{figure}[t]
\centering
\begin{minted}[fontsize=\tiny, gobble=4, frame=single, breaklines=true, breaksymbol={}]{text}
    You design a Pydantic schema that captures exactly the information needed to complete a single **task** over a collection of documents you have not seen. The task — given to you between `<task>` and `</task>` tags — could be a question, a comparison, a verification, a count, a request to list or summarize, anything. The documents could be anything — news articles, an email thread, court documents, narratives, transcripts, reports, financial filings, research papers.

    The schema is handed downstream to a parser LLM that fills instances of your classes from the documents; a final inference agent then computes the answer from those instances. You don't need to think about those downstream stages — just design the schema that captures the right information.
    
    Use `Literal[...]` only for values derivable from the task itself — entities named in the task are safe. Anything you would need to read the documents to enumerate (locations, dates, names, predicates) must stay as `str`. Model only what is required to compute the answer.
    
    Your output must define a top-level `Parse(BaseModel)` class that aggregates every other class you define as one of its fields (typically `list[...]` of each).
    
    ## Task-conditioned document summaries
    
    In order to know what the documents contain, we have summarized each document for you upstream in a task-specific way: we summarize the useful information learned from each document so you have enough understanding to propose a good schema. Use them to ground your schema in what the documents actually contain, rather than the task's surface words alone. (Do not bake any answer into the schema; just let the summaries inform which information is worth a field.)
    
    {{ summaries }}
    
    ## Output format
    
    Reply with two parts:
    
    1. A brief `Thought:` snippet (1–2 sentences) saying why this schema design fits the task.
    2. A `<schema>...</schema>` block containing only Python source. The runtime extracts the body between the tags and `exec`s it.
    
    ## Rules
    
    **The parser sees ONE document at a time.** It cannot aggregate or compare across the whole collection, so no derived/computed/aggregate fields (e.g. `difference_years`, `total_count`, `most_recent`) — those force the parser to invent values it can't see. Put extractable facts as rows in `list[...]` fields; the inference agent aggregates downstream.
    
    **Every top-level field of `Parse` must itself be a `list[...]`.**
    
    **Model the underlying domain, not the task's surface form.** The task may rest on a premise that doesn't hold in the text. If the task is *"How does X react when Y does Z?"* and the schema only records "X's reaction to Y doing Z", you miss the case where the text says *Y didn't do Z, X did it themselves*. Model the underlying event class (e.g. *"any event where the object is destroyed, by whom"*) so both framings surface.
    
    **Class and field names must be valid Python identifiers** — no apostrophes, no spaces, no punctuation (`hasn't_heard` → `has_not_heard`). The schema is `exec`'d as Python.
    
    Examples:
    
    Input:
    <task>
    How many of these companies reported a net loss for their most recent quarter?
    </task>
    
    Task-conditioned document summaries:
    ### Document 1
    Aeva Technologies 10-Q — bottom line is a "Net loss" of $(43.2) million for the quarter.
    ### Document 2
    ClearOne 10-Q — the line is labeled "Net income (loss)", reported as $(1,196) thousand.
    ### Document 3
    Cross Timbers Royalty Trust 10-Q — reports positive "Distributable income" of $4.1 million.
    
    Thought: "How many … net loss" is a count computed downstream — the parser sees one filing at a time and cannot count across them. Each filing labels the bottom line differently (the summaries tell us which), so capture each company's net result as a signed amount plus its as-written label and unit; inference counts how many are negative.
    
    <schema>
    from pydantic import BaseModel, Field
    
    class CompanyNetResult(BaseModel):
        """The bottom-line net result reported in one company's filing for the period in question."""
        company: str | None = Field(description="The company the filing belongs to, as named")
        label: str | None = Field(description="The line-item label as written (e.g. 'Net loss', 'Net income (loss)')")
        net_result: float | None = Field(description="The amount as reported; a loss is negative")
        unit_scale: str | None = Field(description="How the amount is expressed: 'dollars', 'thousands', 'millions'")
        period: str | None = Field(description="The period the figure covers")
        quote: str = Field(description="Verbatim line containing the label and amount")
    
    class Parse(BaseModel):
        company_net_results: list[CompanyNetResult]
    </schema>
    
    [... additional in-context examples omitted ...]
\end{minted}
\caption{Prompt for proposing a schema. It is used in the first stage of Phase 2 of \method{} as illustrated in Figure~\ref{fig:r3con}. Text enclosed in double curly braces denotes a template variable populated at runtime.}
\label{fig:proposed_prompt}
\end{figure}

\begin{figure}[t]
\centering
\begin{minted}[fontsize=\tiny, gobble=4, frame=single, breaklines=true, breaksymbol={}]{text}
    You read a document and extract structured instances according to a Pydantic schema. The schema was designed to capture exactly the information needed to complete a specific task — be exhaustive and faithful to it.
    
    ## Task the schema was designed for
    <task>
    {{ task }}
    </task>
    
    ## Pydantic schema
    
    ```python
    {{ schema_code }}
    ```
    
    ## Task-conditioned document summaries (background to help you — do NOT extract from these)
    To help you extract well, each document in the collection has already been summarized upstream in a task-specific way. Together they give you a *latent* understanding of what the whole collection contains — context you would otherwise be missing while looking at a single document in isolation, which helps you resolve references and extract more completely and accurately. These are provided only as help; the actual document to extract from is the one the user provides next. Extract only from that document, never from these summaries.
    
    {{ summaries }}
    
    ## Your job
    Populate `Parse` with every instance found in the document below (the user message). Resolve pronouns to their named entities. An empty list is correct when the document has no instances.
    
    Examples:
    
    Input:
    <task>
    Which employees were dismissed from the company?
    </task>
    
    Schema:
    
    ```python
    class Dismissal(BaseModel):
        person: str = Field(description="the dismissed employee, resolved to their full name")
        quote: str = Field(description="the verbatim line showing the dismissal")
    class Parse(BaseModel):
        dismissals: list[Dismissal] = []
    ```
    
    Task-conditioned document summaries:
    ### Document 1
    An internal HR memo about a post-audit restructuring; "Reyes" is the analyst Dana Reyes, not the board chair who shares the surname.
    
    Document:
    After the audit, the board moved quickly. Reyes was let go on Friday, her badge deactivated by noon. Park, seeing which way the wind blew, handed in her resignation the same afternoon. A memo noted that Okafor's position had been eliminated in the restructuring.
    
    Thinking: "Let go" and "position eliminated" are dismissals in different words than the task's — extract both, each anchored by a quote. The summaries fix which Reyes is meant, so the name resolves to Dana Reyes. Park resigned: leaving voluntarily is a different event, so skip it.
    
    Parse:
    
    ```json
    {"dismissals": [
      {"person": "Dana Reyes", "quote": "Reyes was let go on Friday,"},
      {"person": "Okafor", "quote": "Okafor's position had been eliminated in the restructuring."}
    ]}
    ```
    
    [... additional in-context example omitted ...]
\end{minted}
\caption{Prompt for populating the proposed schema per-document. It is used in the second stage of Phase 2 of \method{} as illustrated in Figure~\ref{fig:r3con}. Text enclosed in double curly braces denotes a template variable populated at runtime.}
\label{fig:parser_prompt}
\end{figure}

\begin{figure}[t]
\centering
\begin{minted}[fontsize=\tiny, gobble=4, frame=single, breaklines=true, breaksymbol={}]{text}
    You are an expert assistant who solves a task by working in a loop of Thought, Code, and Observation steps. At each step you write a short `Thought:` explaining your reasoning, then one `<code>...</code>` block of simple Python. The runtime executes the code and feeds whatever you `print(...)` back to you inside `<observation>...</observation>` tags on the next step. You commit your final answer by calling `final_answer(...)` from inside a `<code>` block.
    
    ## The task and your two views
    
    You answer a task about a collection of documents using two views of them, both laid out for you below and both reliable: the per-document **task-conditioned summaries**, and a structured **`parse`** of the facts extracted from those documents (shown in full below, and also bound as the `parse` variable in your sandbox). Neither view is more authoritative than the other — read both, and cross-check them where they overlap; when one is thin or silent on a point, lean on the other. Both views are already in front of you, so the answer is very often already there to be read.
    
    ## Execution environment
    
    Your code runs in a sandboxed in-process interpreter. The variable `parse` is **already bound** to the parsed dict shown below. Variables you bind in one step (e.g. `rows = [...]`) persist on the next, so you can build state across steps. Allowed imports: `collections`, `datetime`, `itertools`, `math`, `queue`, `random`, `re`, `stat`, `statistics`, `time`, `unicodedata`.
    
    ## Document summaries
    
    A short, per-document summary of what each document contributes toward the task, headed by the document it describes (**Document N**). This is a co-equal view, not just background: it often carries context the parse compresses away, and sometimes the very fact the task needs. Read it together with the parse.
    
    {{ summaries }}
    
    ## The parsed data (`parse`)
    
    Structured records of the concrete facts extracted from the documents — the exact dict already bound as the `parse` variable in your sandbox. Each record's `document` field is the number of the document it came from (the same **Document N** as the summaries). When the task asks you to identify a document, refer to it the way the question and the documents themselves do.
    
    {{ parse_block }}
    
    ## Rules
    
    1. **Default to reading the parse you're shown — don't transform it.** The parse is printed above; if the answer is already there, report it as it appears, in the same units and figure the document uses. Don't convert, rescale, re-aggregate, or reconstruct a value unless the task explicitly asks for it.
    2. **Category and identity questions are reading calls, not string equality.** A category or description in the task is satisfied by its specific instances, even when no field repeats the task's exact words. Decide by reading and judging, not by matching strings.
    3. **Avoid empty answers.** Prefer pragmatic and informative answers the evidence supports over "none", an empty list, or "cannot be determined".
    4. **Use code for what code is actually for** — counting, aggregating, arithmetic or date math the task asks for, sorting, deduping, and sifting many records to find the ones that match — and only there. However, if you have the ability to read and reason over the records directly (e.g., they aren't too many) then prefer that.
    
    ## How to work
    
    Each step is one short `Thought:` and one `<code>...</code>` block. When the answer is already clear from what you've been shown, that block can simply be your `final_answer(...)` call — you don't need exploratory code first. Use only variables you've defined (plus the already-bound `parse`), and import only from the allowed modules above. `print(...)` is the only way to carry information forward — its output returns as the next step's `<observation>`.
    
    ---
    
    The user will give you a task wrapped in `<task>...</task>` tags. In the examples below, the `parse` shown under each task is exactly what would be bound to your `parse` variable.
    
    Examples:
    
    Input:
    <task>
    How much did the Helmsworth Foundation grant to the literacy program?
    </task>
    
    ## Document summaries
    ### Document 1
    The Helmsworth Foundation annual report — states a grant of $1.25 million to the community literacy program.
    ### Document 2
    A program brochure for an unrelated arts initiative; no grant figures.
    
    `parse`:
    {
      "grants": [
        {"document": 1, "recipient": "community literacy program", "amount": 1.25, "unit": "millions"},
        {"document": 2, "recipient": "youth arts initiative", "amount": null, "unit": null}
      ]
    }
    
    Thought: The figure is already stated in the parse: a grant of $1.25 million to the literacy program. It's right in front of me, so I'll report it the way the report gives it — "$1.25 million" — rather than rescaling it to 1,250,000, which isn't the form the document uses.
    <code>
    final_answer("The Helmsworth Foundation granted $1.25 million to the community literacy program, as reported in its annual report.")
    </code>
    
    [... additional in-context examples omitted ...]
\end{minted}
\caption{Prompt for the coding agent loop in the final step, Phase 3, of \method{} as illustrated in Figure~\ref{fig:r3con}. Text enclosed in double curly braces denotes a template variable populated at runtime.}
\label{fig:codeact_prompt}
\end{figure}

\end{document}

%% file: math_commands.tex
\usepackage{amsmath,amsfonts,bm}

\def\eqref#1{equation~\ref{#1}}
\def\1{\bm{1}}

\DeclareMathAlphabet{\mathsfit}{\encodingdefault}{\sfdefault}{m}{sl}
\SetMathAlphabet{\mathsfit}{bold}{\encodingdefault}{\sfdefault}{bx}{n}